\PassOptionsToPackage{table,dvipsnames}{xcolor}
\documentclass[]{selfevolagent}
\usepackage{microtype}
\usepackage{amsfonts}
\usepackage{xcolor}
\definecolor{lavender}{RGB}{162, 185, 250}
\definecolor{softlavender}{RGB}{216, 226, 251}
\definecolor{selfevolagent}{RGB}{117, 150, 248} 
\usepackage{graphicx}
\usepackage{booktabs}
\usepackage{wrapfig}
\usepackage{siunitx} 
\usepackage{float}
\usepackage{amsmath}
\usepackage{amsthm}
\usepackage{subcaption}
\usepackage{makecell}
\usepackage{algpseudocode}
\usepackage[linesnumbered,lined,boxed,commentsnumbered,ruled,longend]{algorithm2e}
\theoremstyle{plain}

\theoremstyle{definition}

\theoremstyle{remark}

\usepackage{enumitem}
\usepackage{pifont}
\usepackage[edges]{forest}
\usetikzlibrary{arrows.meta}
\usepackage{tikz}
\usepackage{graphicx}     
\usepackage{booktabs}     
\usepackage{enumitem}     
\usepackage[most]{tcolorbox} 
\usepackage{fontawesome5}
\usepackage{booktabs} 
\usepackage{url}
\usepackage{wrapfig}
\usepackage{longtable}
\usepackage{tabularx}
\usetikzlibrary{
  positioning,
  calc,
  shapes.symbols,
  shapes.geometric,
  shapes.misc
}
\usepackage[T1]{fontenc}
\definecolor{selfevolagent_dark}{HTML}{1B42E0}   
\definecolor{selfevolagent_light}{HTML}{5276F6}  
\definecolor{selfevolagent_lighter}{HTML}{EDF1FA}

\usepackage{listings}
\usepackage{xcolor}

\definecolor{codebg}{RGB}{245,245,245}
\definecolor{codecomment}{RGB}{0,128,0}
\definecolor{codekeyword}{RGB}{220,20,60}

\lstdefinestyle{pytorchstyle}{
    backgroundcolor=\color{codebg},   
    commentstyle=\color{codecomment},
    keywordstyle=\color{codekeyword},
    numberstyle=\tiny\color{gray},
    basicstyle=\ttfamily\footnotesize,
    breaklines=true,                 
    keepspaces=true,                 
    numbers=left,                    
    numbersep=5pt,                  
    showspaces=false,                
    showstringspaces=false,
    tabsize=4,
    language=Python
}

\usepackage{svg}

\usepackage{xltabular}
\usepackage{booktabs}
\usepackage{caption}

\usepackage{fontawesome5}

\newcommand{\ghlink}[1]{\faIcon{github}\,\href{#1}{GitHub}}
\newcommand{\weblink}[1]{\faIcon{globe}\,\href{#1}{Website}}

\definecolor{rootcolor}{RGB}{27, 66, 224}
\definecolor{catcolor}{RGB}{82, 118, 246} 
\definecolor{subcatcolor}{RGB}{145, 172, 250} 
\definecolor{papercolor}{RGB}{237, 241, 250}

\title{\centering EvoEmbedding: Evolvable Representations for Long-Context Retrieval and Agentic Memory}
\author{Chang Nie}
\author{Chaoyou Fu$^{\ddagger}$}
\author{Junlan Feng}
\author{Caifeng Shan}
\affiliation{
Nanjing University
}

\abstract{
\begin{abstract}

Existing embedding models are inherently static: they encode text segments in isolation, ignoring their surrounding context and temporal order.
This paper introduces \textbf{EvoEmbedding}, a novel embedding model that generates \textit{evolvable} representations for retrieval.
It is tailored for long-context scenarios, where information is dynamic, sequential, and requires continuous state tracking.
Our design is simple: EvoEmbedding maintains a continuously updated latent memory as it sequentially processes inputs, and uses it alongside the raw content to jointly generate evolvable embeddings.
Consequently, for the same query, our model adapts its representation to retrieve distinct targets based on the evolving context, going beyond static semantic search.
To equip the model with this capability, we construct \textbf{EvoTrain-180K}, a diverse dataset for the joint optimization of latent memory and retrieval.
Furthermore, we introduce a memory queue to prevent representation collapse during recurrent encoding, alongside segment-batching techniques that tackle significant length variance and accelerate training by 3.8$\times$.
Extensive experiments show that our model not only outperforms larger-scale specialists (e.g., Qwen3-Embedding-8B and KaLM-Embedding-Gemma3-12B) across a range of long-context retrieval benchmarks, but also generalizes well to downstream tasks (e.g., personalization) with contexts 10$\times$ longer than its training window.
Notably, EvoEmbedding seamlessly integrates into agentic workflows to boost performance. For instance, a naive RAG pipeline equipped with our model surpasses dedicated agentic memory systems.

\end{abstract}
}
\metadata[\faEnvelope\ Main Contact]{changnie@smail.nju.edu.cn, bradyfu24@gmail.com}
\metadata[{\raisebox{-0.2ex}{\includegraphics[height=1em]{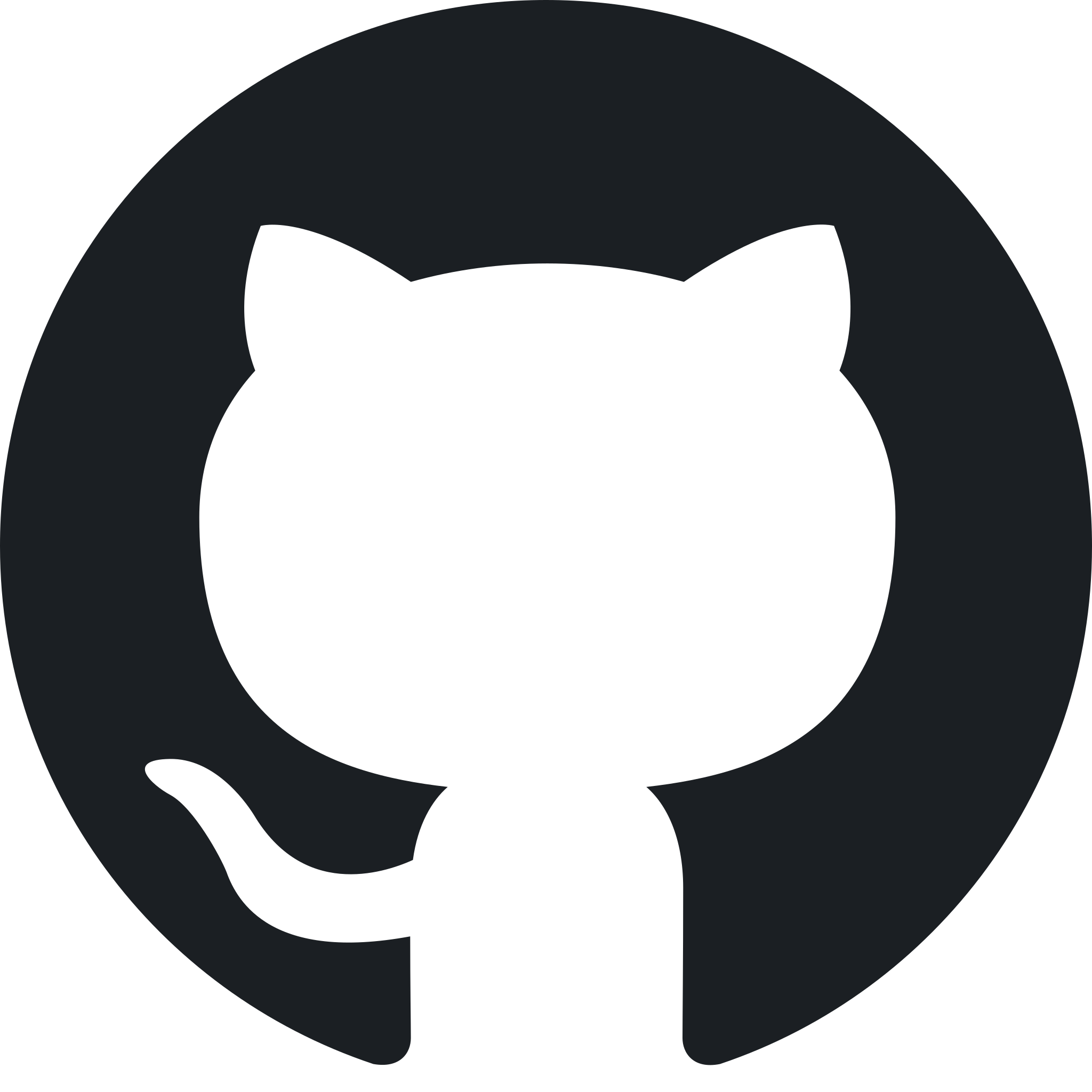}}\ Github}]{\url{https://clare-nie.github.io/EvoEmbedding}}

\begin{document}
 \maketitle
\begin{figure}[!h]
    \centering
    \includegraphics[width=0.9\linewidth]{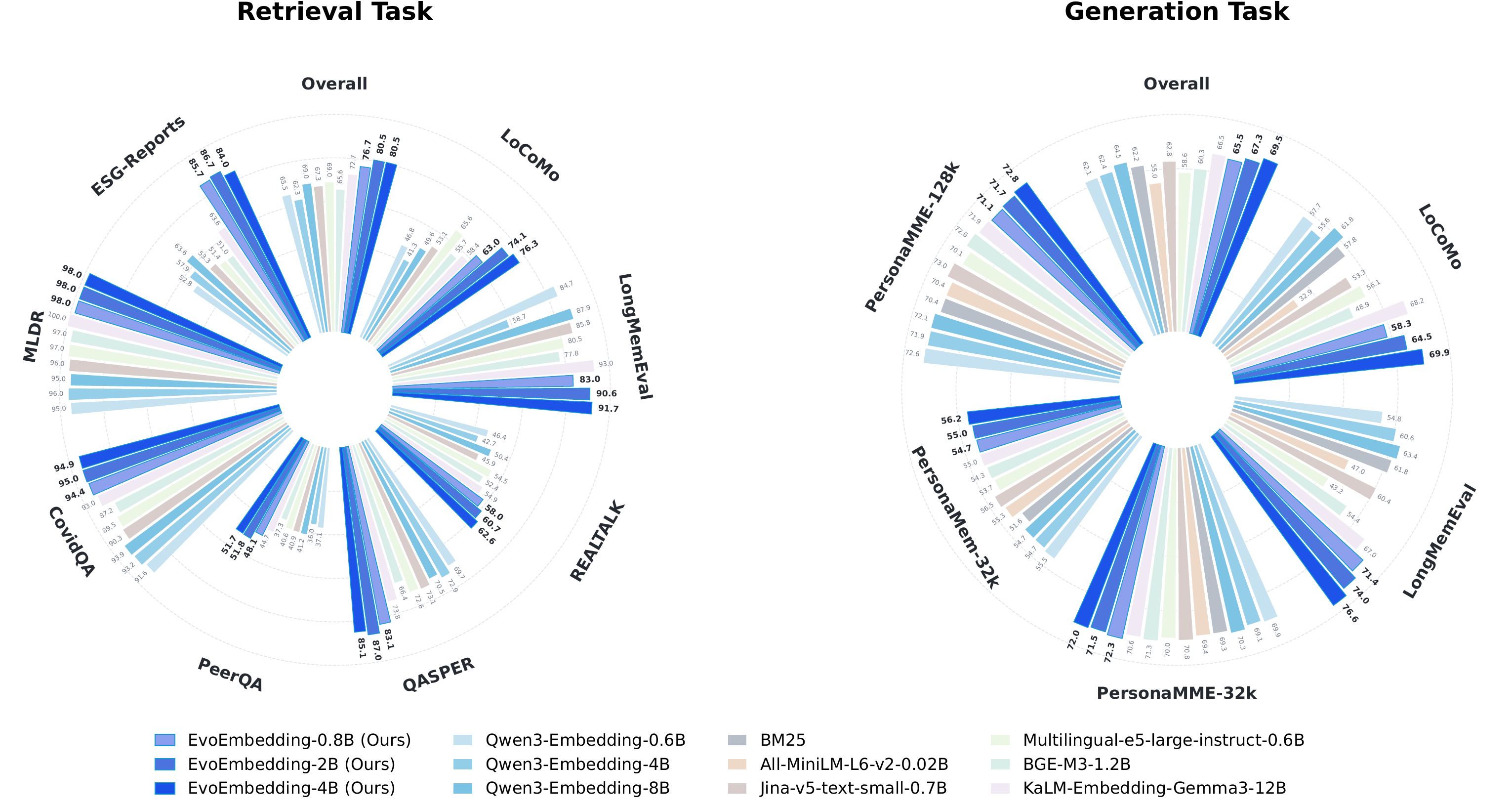}
\caption{\textbf{Performance comparison across long-context retrieval and generation tasks.} Our \textcolor[HTML]{5E83E7}{\textbf{EvoEmbedding}} family (0.8B, 2B, and 4B) achieves superior results across 10 benchmarks.
Notably, our model excels in handling dynamic contexts (e.g., long-term personalization), outperforming both established static and large-scale models.
We report Recall@10 for retrieval, and generation accuracy using a naive RAG pipeline based on the Top-4 retrieved segments.}
    \label{fig:intro-main}
\end{figure}

\section{Introduction}
\label{intro}

Retrieval-Augmented Generation (RAG) has emerged as a dominant paradigm for augmenting Large Language Models (LLMs) with external knowledge and extended context~\citep{fan2024survey,zhao2026retrieval}.
It is widely applied in AI agent design and context engineering~\citep{mei2025survey,openclaw2026}, equipping LLMs with essential long-term memory capabilities to accomplish complex, long-horizon tasks~\citep{hu2025memory}.
However, when confronted with long-context scenarios where information is dynamic, sequential, and requires continuous state tracking, existing retrieval systems, which predominantly rely on static representations, frequently fail to retrieve the desired target information~\citep{maharana2024evaluating,weller2025theoretical}, and often introduce noise into the generation process~\citep{liu2024lost}.

\begin{figure}[!t]
\centering
\includegraphics[width=1.0\linewidth]{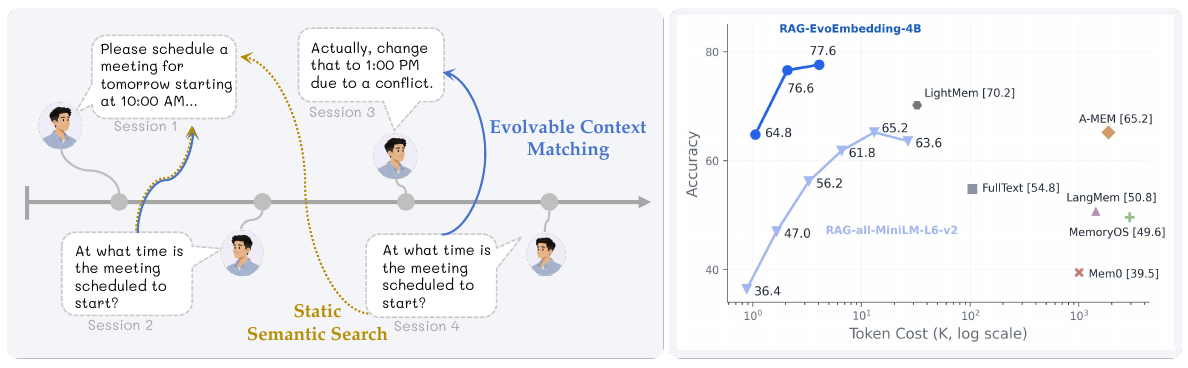}
\caption{\textbf{(Left)} For the same query under different contexts, EvoEmbedding generates evolvable representations that avoid the outdated retrievals of static methods. \textbf{(Right)} On LongMemEval, standard RAG equipped with EvoEmbedding-4B outperforms existing memory baselines, achieving SOTA performance while minimizing costs with \textit{zero} token overhead for explicit memory construction.}
    \label{fig:intro}
\end{figure}

The root cause of this limitation lies in two key flaws of current retrieval pipelines. First, existing methods typically extract and encode text segments in isolation~\citep{gao2023retrieval, sarthi2024raptor}, a process that inherently disrupts the temporal continuity and contextual associations between these segments.
Second, current embedding models are optimized via contrastive learning primarily on short, static samples~\citep{chen2024bge,zhang2025qwen3,zhao2025kalm}. This restricts their capability to the mere discrimination of semantic relevance, rendering them ill-equipped for complex contextual understanding, such as coreference resolution and temporal reasoning~\citep{guo2024lightrag,weller2025theoretical}.
As illustrated in Fig.~\ref{fig:intro} (Left), a user initially schedules a meeting and later postpones it.
When queried about the meeting time under varying contexts, the static embedding model fails to capture this dynamic evolution, ultimately retrieving the outdated schedule information.

To mitigate these limitations, recent advancements primarily optimize the RAG pipeline across two aspects: indexing and retrieval execution.
During the indexing phase, efforts focus on transforming historical contexts into retrieval-friendly databases.
This involves performing coreference resolution~\citep{liu2026simplemem}, constructing structured memories~\citep{chhikara2025mem0}, and 
augmenting raw texts with generative metadata like captions, session summaries, and keywords~\citep{nie2026personavlm}.
During the retrieval phase, existing systems seek to bridge the semantic gap by optimizing query representations through rewriting~\citep{ma2023query}, and frequently introduce additional reranker models to refine the candidate list~\citep{li2026query}.
Furthermore, the emerging paradigm of agentic RAG~\citep{singh2025agentic} integrates autonomous agents to orchestrate dynamic query routing and multi-step iterative retrieval, aiming to compensate for the inherent limitations of static embeddings through sophisticated workflow orchestration~\citep{nie2026personavlm,liu2026simplemem}.
Although these approaches alleviate certain issues, they incur substantial computational overhead, increase inference latency, and remain suboptimal.

This paper introduces \textbf{EvoEmbedding}, which overcomes the flaws of static models at the representation level by generating contextually \textit{evolvable} embeddings. 
Specifically, it maintains a fixed-capacity latent memory to preserve evolving contextual states.
Upon receiving new inputs (e.g., streaming text chunks or dialogue turns), it updates this latent memory and generates context-aware embeddings in parallel.
This design allows EvoEmbedding to recurrently compress ongoing contexts, injecting rich temporal dynamics and cross-segment associations into the representations.
To facilitate training, we construct \textbf{EvoTrain-180K}, a diverse dataset tailored for the joint optimization of memory and retrieval capabilities.
Given the wide variation in context lengths within this dataset, we propose the memory queue and segment-batching techniques. These techniques successfully prevent representation collapse and improve training efficiency by 3.8$\times$ without curriculum learning~\citep{bulatov2023scaling}.
Finally, we employ lightweight LoRA adaptation to equip general-purpose LLMs with strong representation capabilities while avoiding catastrophic forgetting.
By applying this unified architecture across different base models, we develop the EvoEmbedding family, scaled at 0.8B, 2B, and 4B\footnote{For brevity, unless otherwise specified, EvoEmbedding denotes our flagship 4B variant throughout this paper.} parameters for broad applicability.

Extensive experiments across 10 benchmarks encompassing diverse tasks, domains, and context scales demonstrate that EvoEmbedding delivers:
(i) \textbf{State-of-the-art retrieval performance}.
As illustrated in Fig.~\ref{fig:intro-main}, EvoEmbedding yields the highest overall accuracy across eight long-context benchmarks.
It surpasses generalist KaLM-Embedding-Gemma3-12B~\citep{zhao2025kalm} by +6.4\% and Qwen3-Embedding-8B~\citep{zhang2025qwen3} by +11.1\%.
(ii) \textbf{Strong scalability and generalization}.
EvoEmbedding generalizes to downstream long-context understanding and personalization tasks, effectively handling 128K contexts (10$\times$ its maximum training window, and >100$\times$ its average sample length of 1.2K).
Furthermore, as depicted in Fig.~\ref{fig:intro} (Right), naive RAG with EvoEmbedding-4B achieves the highest accuracy with minimal token consumption on LongMemEval-s~\citep{wu2024longmemeval}, substantially outperforming memory-based methods (77.6\% vs. A-MEM's 65.2\% and LightMem's 70.2\%)~\citep{fang2025lightmem,xu2026mem}.
(iii) \textbf{Enhancement of agentic memory systems}.
EvoEmbedding seamlessly integrates into existing agentic memory systems and delivers substantial gains (e.g., {+19.2\%} for A-MEM and {+13.5\%} for LightMem), consistently outperforming strong baselines such as Qwen3-Reranker-4B and complex reasoning-based retrieval strategies.
(iv) \textbf{Temporal retrieval capabilities}.
We show our model is inherently suited for temporal RAG. When confronted with temporal keywords such as `\textit{firstly}' or `\textit{lastly}', the query-context similarities peak distinctly at the targeted historical stages, successfully decoupling temporal intents from coarse textual semantics.
Overall, these compelling advantages highlight the potential of EvoEmbedding to serve as the next-generation embedding model for long-context retrieval.

\section{Related Work}
\label{rel}

\textbf{Text Embedding Models.}
Text embedding models are widely applied in natural language processing (NLP) tasks such as retrieval and reranking~\citep{zhao2025kalm}.
A representative early milestone is Sentence-BERT~\citep{reimers2019sentence}, which encodes variable-length text into dense semantic vectors and evaluates text relevance via similarity.
Recently, leveraging LLMs as embedding models has emerged as the dominant trend.
By repurposing generative LLMs for representation learning, pioneering models like E5~\citep{wang2024multilingual}, Qwen3-Embedding~\citep{zhang2025qwen3}, and KaLM-Embedding~\citep{zhao2025kalm} have achieved remarkable performance across diverse retrieval benchmarks~\citep{muennighoff2023mteb}.
When confronted with extended contexts, the standard practice is to segment the long input into fixed-size chunks and encode them independently.
However, this static and fragmented representation severely disrupts the temporal continuity and contextual associations between segments, frequently leading to imprecision and redundancy during the subsequent retrieval process~\citep{gao2023retrieval, cuconasu2024power}.

\textbf{Agentic RAG for Long-Context.}
To overcome the limitations of static embeddings, agentic RAG optimizes the retrieval pipeline during information storage and query execution~\citep{singh2025agentic}.
On the one hand, existing works strive to construct retrieval-friendly databases~\citep{chhikara2025mem0,fang2025lightmem}, introduce auxiliary storage structures such as knowledge graphs~\citep{han2024retrieval,guo2024lightrag} and various memory architectures~\citep{hu2025memory}, and integrate post-processing modules like rerankers~\citep{zhang2025qwen3} to boost retrieval performance.
On the other hand, frontier approaches leverage autonomous agents to orchestrate the querying process, introducing techniques such as query rewriting~\citep{ma2023query}, iterative reasoning~\citep{nie2026personavlm}, and multi-agent collaboration~\citep{nguyen2025ma} to balance computational overhead and retrieval precision.
Despite these advancements, such complex pipelines remain suboptimal due to the inherent loss of global context, and incur prohibitive latency that hinders their application in real-time and responsive scenarios.

\textbf{Latent Memory and Retrieval.}
Early studies have explored maintaining long-term memory internally to bypass external retrieval altogether.
For instance, the Recurrent Memory Transformer (RMT)\citep{bulatov2022recurrent,bulatov2023scaling} extends the context window to over 1M tokens, while maintaining linear computational complexity by continuously updating a latent memory in place.
To scale up latent memory, M+~\citep{wang2025m} employs a co-trained retriever to dynamically recall compressed hidden states during generation.
Furthermore, LatentRAG~\citep{zheng2026latentrag} and LAnR~\citep{nguyen2026latent} explore a more integrated paradigm, jointly performing reasoning, retrieval, and generation within the model's latent space. While these approaches demonstrate the potential of latent representations, existing mechanisms remain deeply coupled with the autoregressive generation process of LLMs. This architectural entanglement severely limits their practical applicability, as they require full white-box access to internal hidden states, rendering them incompatible with closed-source commercial models accessed via APIs.
Moreover, forcing a single model to simultaneously manage latent memory and token generation without explicit isolation can lead to unpredictable behaviors such as hallucinations~\citep{xu2026more}.

\section{Methodology}
\label{evo}

\begin{figure}[!t]
\centering
\includegraphics[width=1.0\linewidth]{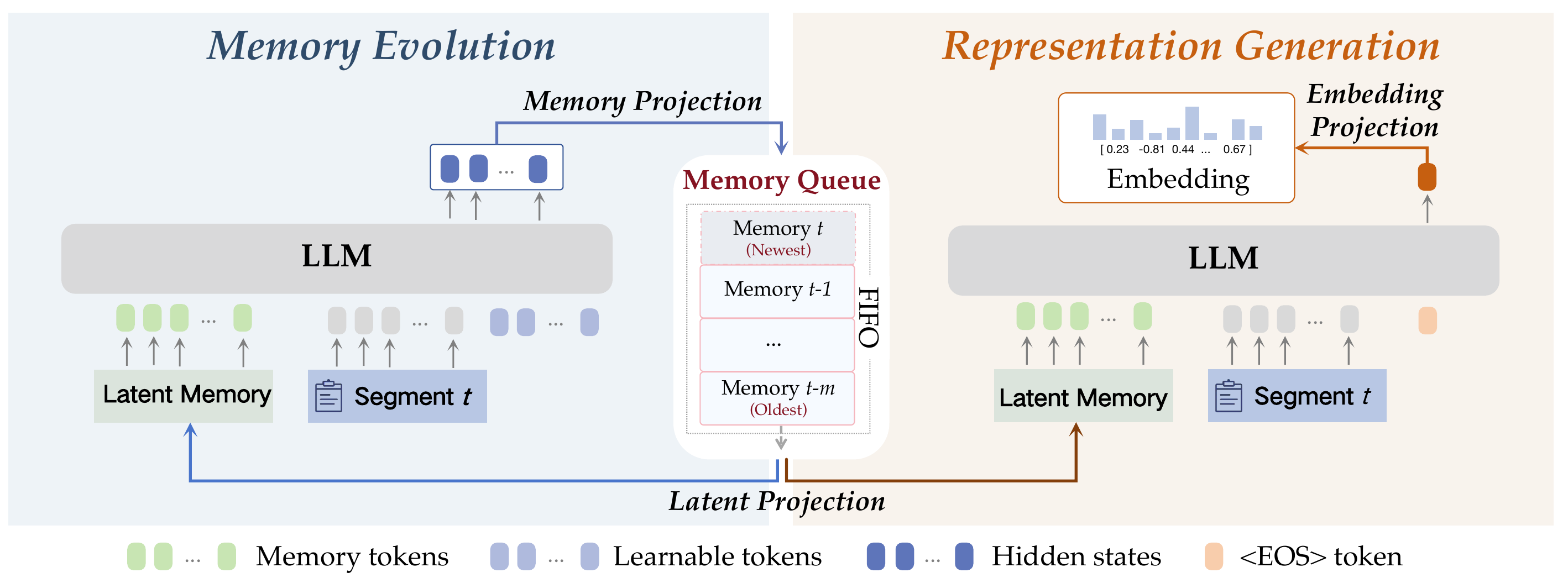}
\caption{\textbf{Overview of the proposed EvoEmbedding.} The model performs two parallel tasks for the input segment at step $t$. {(Left)} \textbf{Memory Evolution}: The LLM compresses the segment and integrates the previous memory into learnable tokens, which are projected to update a FIFO latent memory queue. {(Right)} \textbf{Representation Generation}: The historical latent memory is combined with the current segment to generate a contextually evolvable embedding for retrieval.}
\label{fig:fra}
\end{figure}

\subsection{EvoEmbedding}
\textbf{Overview:}
EvoEmbedding sequentially processes segments\footnote{A segment refers to a text unit to be retrieved, such as a single sentence or a conversational turn.} split from a long input sequence.
As illustrated in Fig.~\ref{fig:fra}, given the current input segment $x_t$ and the latent memory $\mathbf{M_{t-1}}$, EvoEmbedding performs memory evolution and representation generation tasks in parallel.
This process can be formulated as:
\begin{equation}
\begin{aligned}
\mathbf{\tilde{M}_t} &= \pi_{\theta_m}(x_t, \mathbf{M_{t-1}}),\\
\mathbf{v_t} &= \pi_{\theta_r}(x_t, \mathbf{M_{t-1}}),
\end{aligned}
\label{eq1}
\end{equation}
where $\theta_m$ and $\theta_r$ denote the task-relevant parameters of the model $\pi$,
$\mathbf{\tilde{M}_t} \in \mathbb{R}^{K \times D}$ represents the newly generated $K$ latent tokens,
and $\mathbf{v_t} \in \mathbb{R}^{D_{emb}}$ is the corresponding vector of $x_t$.
During the query phase, only the representation generation task is executed to obtain the corresponding retrieval vector. 
This formulation mirrors the core design of EvoEmbedding: it dynamically maintains a latent memory as the global semantic context and jointly encodes this context with the current input to produce evolvable embeddings.

\noindent \textbf{Latent Memory Queue.} 
We design a queue mechanism to manage and update the latent memory of EvoEmbedding.
Specifically, the updated memory $\mathbf{M_t}$ is constructed as:
\begin{equation}
\mathbf{M_t}= \text{Queue}(\mathbf{M_{t-1}},f_m(\mathbf{\tilde{M}_t})),
\label{eq2}
\end{equation}
where $\mathbf{M}_t \in \mathbb{R}^{C \times D}$ is a First-In-First-Out (FIFO) queue matrix. The memory capacity is defined as $C = L \times K$, designed to store the latent representations generated from the most recent $L$ steps.
$f_m(\cdot)$ is a projector
to map the newly generated tokens $\mathbf{\tilde{M}_t}$ into the shared memory space.
The rationale behind this queue-based design is twofold:
(i) \textbf{Bounded loop:} It guarantees that a single historical memory is loop-encoded at most $L$ times. This avoids the collapse caused by recurrent encoding of memory~\citep{yu2026latent}, allowing EvoEmbedding to be directly trained on long contexts without the need for curriculum learning (see Table~\ref{tab:ablation_results}).
(ii) \textbf{Bounded Capacity:} By strictly limiting the memory size, it bounds the computational complexity and explicitly forces the model to learn to fuse new knowledge with historical states at each step.
Compared to M$+$~\citep{wang2025m}, which uses cached layer-wise features as memory, we only store $C=512$ latent tokens, whose memory footprint is roughly equivalent to that of an encoded image.

\noindent \textbf{Dynamic Segment-Batching.}
We introduce a dynamic Segment-Batching technique to overcome the efficiency challenges of processing segment-level inputs sequentially during both training and inference.
Instead of executing the forward pass segment-by-segment, EvoEmbedding processes $k$ consecutive segments in parallel.
We dynamically determine $k$ to ensure that the total length of the concatenated inputs does not exceed a predefined threshold (e.g., $2048$ tokens). 
Accordingly, the memory evolution can be formulated as the batched form of Eq.~(\ref{eq1}), calculated by
$\mathbf{\tilde{M}_{t:t+k}} = \pi_{\theta_m}(x_{t:t+k}, \mathbf{M_{t-1}})$.
The subsequent memory queue updates and embedding generation processes remain consistent with the previous formulation.
This design not only achieves a 3.8$\times$ speedup but also improves performance (see Table~\ref{tab:ablation_results}).

\subsection{Training and Optimization}
EvoEmbedding is initialized from general language models, such as the Qwen series~\citep{yang2025qwen3,qwen35blog}, allowing us to build a model family scaled at 0.8B, 2B, and 4B parameters. 
We employ a multi-LoRA~\citep{hu2022lora} design to decouple the memory evolution and representation generation capabilities, allowing for flexible switching during inference.
Given an input sample consisting of $t$ segments $\{x_i\}_{i=1}^{t}$ and a query $q$, we first construct the latent memory $\mathbf{M_t}$ and obtain the embeddings $\{\mathbf{v_i}\}_{i=1}^{t}$ and $\mathbf{v_q}$ based on Eq.~(\ref{eq1}).
We then jointly optimize the model using a combined objective:
\begin{equation}
\mathcal{L} = \mathcal{L}_{mem} + \mathcal{L}_{con},
\label{eq3_total}
\end{equation}
where $\mathcal{L}_{mem}$ is the memory generation loss and $\mathcal{L}_{con}$ is the contrastive representation loss.
The first term $\mathcal{L}_{mem}$ is measured using standard cross-entropy. We use the generated latent memory $\mathbf{M}_t$ and the query $q$ as context to predict the target answer $y$:
\begin{equation}
\mathcal{L}_{mem} = - \sum_{j=1}^{|y|} \log P(y_j \mid y_{<j}, q, \mathbf{M}_t).
\label{eq3_mem}
\end{equation}
Crucially, during this prediction step, we keep the backbone LLM parameters frozen and deactivate all LoRA adapters.
This design ensures that the loss backpropagates through the frozen backbone directly into $\mathbf{M}_t$, implicitly forcing the memory module $\theta_m$ to generate latent states that are compatible with the base LLM's native semantic space.

The second term $\mathcal{L}_{con}$ is the contrastive loss designed to align the query representation with the relevant contexts. Unlike standard retrieval settings~\citep{zhang2025qwen3,zhao2025kalm}, our candidate pool is dynamically partitioned from the $t$ segments of the current sample. Let $\mathcal{P} = \{\mathbf{v}_i^+\}_{i=1}^{P}$ denote the set of embeddings for the positive segments containing the supporting evidence, and $\mathcal{N} = \{\mathbf{v}_j^-\}_{j=1}^{N}$ denote the set of embeddings for the negative segments, where $P + N = t$.
To optimize across multiple positive targets and balance the learning difficulty across varying context lengths, we formulate a length-weighted multi-positive contrastive loss:
\begin{equation}
\mathcal{L}_{con} = \frac{\log(N + 1)}{P} \sum_{i=1}^{P} \left( - \log \frac{\exp(\mathbf{v}_q^\top \mathbf{v}_i^+ / \tau)}{\exp(\mathbf{v}_q^\top \mathbf{v}_i^+ / \tau) + \sum_{j=1}^{N} \exp(\mathbf{v}_q^\top \mathbf{v}_j^- / \tau)} \right),
\label{eq3_con}
\end{equation}
where $\mathbf{v}_q = \pi_{\theta_r}(q, \mathbf{M}_T)$ represents the normalized embedding of the query encoded with the final memory state, and $\tau=0.1$ is a temperature hyperparameter. The term $\log(N + 1)$ serves as a length-weighting factor. Since the number of negative distractors $N$ varies significantly with the input length, this factor adaptively calibrates the loss scale, ensuring stable optimization across highly variable sequence lengths.

\begin{figure*}[t]
\centering
\includegraphics[width=\textwidth]{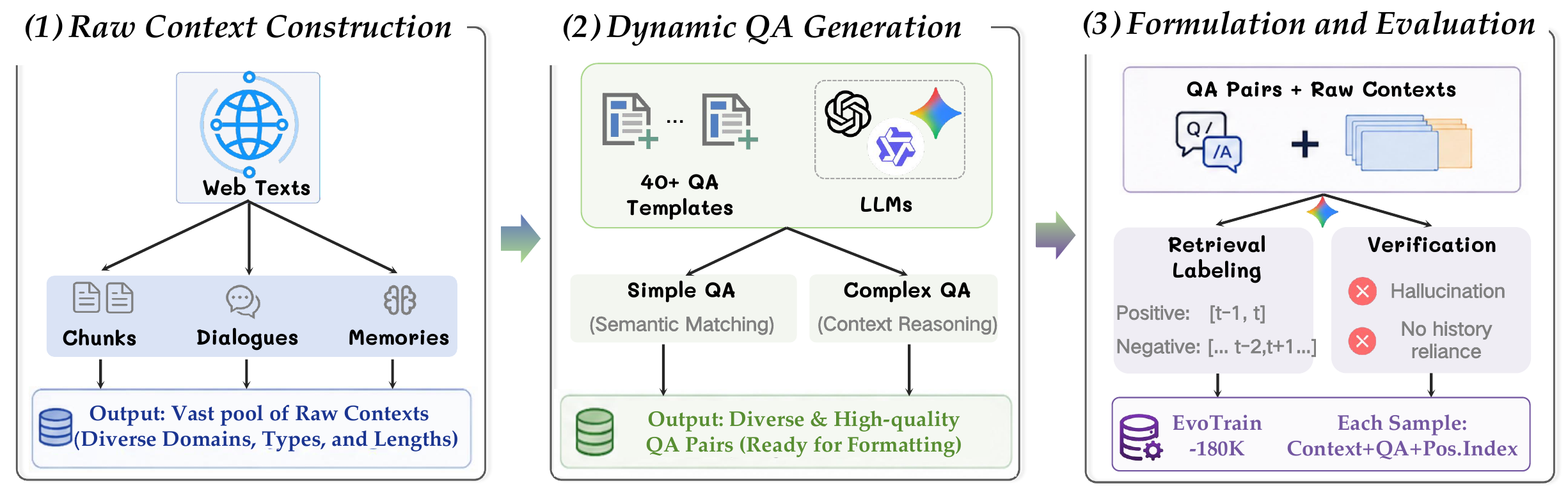}
\caption{\textbf{Construction pipeline of the EvoTrain-180K dataset.} The process comprises three stages: (1) \textbf{Raw Context Construction} across diverse domains, formats, and lengths; (2) \textbf{Dynamic QA Generation} utilizing LLMs and 40+ templates to produce both semantic and reasoning-based queries; and (3) \textbf{Formulation and Verification} to perform positive/negative retrieval labeling and filter out noisy samples (e.g., hallucinations and context-independent queries).}
\label{fig:dataset_pipeline}
\end{figure*}

\subsection{Construction of EvoTrain-180K}
We construct \textbf{EvoTrain-180K} specifically for the training of EvoEmbedding.
It is a large-scale, multi-domain dataset tailored for long-context retrieval.
As shown in Fig.~\ref{fig:dataset_pipeline}, we design an automated three-stage data synthesis pipeline to generate high-quality training samples.

\begin{itemize}
\item \textbf{Stage1: Raw Context Construction.}
We construct three primary types of contexts:
(i) \textit{Diverse web texts}: We randomly sample documents from the FineWeb\footnote{https://huggingface.co/datasets/HuggingFaceFW/fineweb} dataset and process them into sequential segments using a sliding window approach.
(ii) \textit{Dialogues}: We employ powerful LLMs to synthesize multi-turn, persona-driven dialogues based on predefined topics.
(iii) \textit{Memories}: We extract various types of memory from the raw texts or dialogues to serve as context.
This stage yields a vast pool of raw contexts spanning diverse domains, types, and lengths.

\item \textbf{Stage2: Dynamic QA Generation.}
This stage builds QA pairs based on the given contexts and introduces two specific designs to ensure sample diversity:
(i) We pre-define over 40 template types (e.g., coreference resolution, temporal understanding) to guide the QA generation.
(ii) We utilize LLMs of varying types and sizes to generate questions. This ensures the inclusion of both simple questions for basic semantic matching and complex questions for deep context understanding.

\item \textbf{Stage3: Retrieval Formulation and Evaluation.}
This final stage employs the strong $\texttt{Gemini-3.1-Pro-Preview}$\footnote{https://deepmind.google/models/gemini/pro} for retrieval labeling and sample verification.
The labeling process identifies the indices of query-relevant segments to serve as the positive target $v^+$.
The verification process ensures valid QA pairs by ruling out hallucinations and enforcing reliance on provided history over general knowledge.
\end{itemize}
Through this rigorous pipeline, we construct 184,137 high-quality samples to jointly train EvoEmbedding's memory and retrieval capabilities.
To ensure training efficiency, each sample is strictly constrained to a maximum of 12K tokens and 256 segments.
Although trained on less than 1\% of the data volume used by contemporary embedding models~\citep{zhao2025kalm} and with a training context length under one-tenth of that in testing scenarios~\citep{wu2024longmemeval,zhao2026lmeb,nie2026personavlm}, EvoEmbedding demonstrates exceptional scalability and robust generalization.
Further details are provided in Appendix~\ref{app:data}.

\begin{table*}[!t]
\centering
\caption{Retrieval performance across diverse long-context and conversational benchmarks. We report Recall@10 and NDCG@10 following the evaluation protocol~\citep{zhao2026lmeb}. 
The \textbf{EvoEmbedding family} achieves the best overall results across all scales, outperforming strong established baselines (e.g., Qwen3-Embedding-8B and KaLM-12B) with much smaller parameter sizes and embedding dimensions. 
The best results are highlighted in \textbf{bold}, and the second best are \underline{underlined}.}
\label{tab:retrieve_topk_results}
\resizebox{\textwidth}{!}{
\begin{tabular}{l c c l | cccccccc | c}
\toprule
 \textbf{Model} & \textbf{Size} & \textbf{Dim} & \textbf{Metric} & \textbf{ESG-Reports} & \textbf{LoCoMo} & \textbf{LongMemEval} & \textbf{REALTALK} & \textbf{QASPER} & \textbf{PeerQA} & \textbf{CovidQA} & \textbf{MLDR} & \textbf{Overall} \\
\midrule

Jina-v5-text-small & 0.7B & 1024 &R@10&  53.3 & 53.1 & 85.8 & 45.9 & 73.1 & 40.9 & 90.3 & 96.0 & 67.3 \\
Multilingual-e5-large & 0.6B & 1024 & R@10
& 51.4 & 65.6 & 80.5 & 54.5 & 72.6 & 40.6 & 89.5 & 97.0 & 69.0 \\
BGE-M3 &1.2B & 1024 & R@10
& 51.0 & 55.7 & 77.8 & 52.4 & 66.4 & 37.3 & 87.2 & 97.0 & 65.6 \\
KaLM-Embedding-Gemma3 &12B & 3840 & R@10
& 63.6 & 58.4 & \textbf{93.0} & 54.9 & 73.8 & 44.7 & 93.0 & \textbf{100.0} & 72.7 \\
Qwen3-Embedding-0.6B &0.6B & 1024 & R@10
& 52.8 & 46.8 & 84.7 & 46.4 & 69.7 & 37.1 & 91.6 & 95.0 & 65.5 \\
Qwen3-Embedding-4B &4B   & 2560 & R@10
& 57.9 & 41.3 & 58.7 & 42.7 & 72.9 & 36.0 & 93.2 & 96.0 & 62.3 \\
Qwen3-Embedding-8B&8B   & 4096 & R@10
& 63.6 & 49.6 & 87.9 & 50.4 & 70.5 & 41.2 & 93.9 & 95.0 & 69.0 \\
\rowcolor{cyan!15}
\textbf{EvoEmbedding-0.8B (Ours)} & 0.8B & 1024 & R@10 & \underline{85.7} & 63.0 & 83.0 & 58.0 & 83.1 & 48.1 & 94.4 & \underline{98.0} & \underline{76.7} \\
\rowcolor{cyan!15}
\textbf{EvoEmbedding-2B (Ours)}   & 2B   & 1024 & R@10 & \textbf{86.7} & \underline{74.1} & 90.6 & \underline{60.7} & \textbf{87.0} & \textbf{51.8} & \textbf{95.0} & \underline{98.0} & \textbf{80.5}\\
\rowcolor{cyan!15} 
\textbf{EvoEmbedding-4B (Ours)} & 4B & 1024 & R@10
& 84.0 & \textbf{76.3} & \underline{91.7} & \textbf{62.6} & \underline{85.1} & \underline{51.7} & \underline{94.9} & \underline{98.0} & \textbf{80.5} \\
\midrule
\midrule

Jina-v5-text-small &0.7B & 1024 & N@10
& 36.8 & 39.4 & 74.6 & 36.9 & 50.7 & 30.3 & 73.0 & 76.5 & 52.3 \\
Multilingual-e5-large &0.6B & 1024 & N@10
& 35.2 & 50.1 & 69.9 & 44.9 & 51.3 & 32.4 & 69.3 & \underline{80.9} & 54.2 \\
BGE-M3 & 1.2B & 1024 & N@10
& 35.9 & 41.8 & 65.6 & 42.0 & 44.2 & 28.2 & 65.1 & 77.0 & 50.0 \\
KaLM-Embedding-Gemma3 & 12B & 3840 & N@10
& 49.5 & 42.8 & \textbf{81.4} & 44.4 & 51.5 & 32.5 & {78.1} & 76.4 & 57.1 \\
Qwen3-Embedding-0.6B & 0.6B & 1024 & N@10
& 37.4 & 34.2 & 72.8 & 37.2 & 47.3 & 27.3 & 74.2 & 76.6 & 50.9 \\
Qwen3-Embedding-4B &4B   & 2560 & N@10
& 40.8 & 29.9 & 47.1 & 34.5 & 50.2 & 27.2 & 76.7 & 77.8 & 48.0 \\
Qwen3-Embedding-8B & 8B   & 4096 & N@10
& 43.2 & 36.4 & 77.6 & 41.2 & 47.4 & 30.1 & \textbf{79.4} & 77.5 & 54.1 \\
\rowcolor{cyan!15}
\textbf{EvoEmbedding-0.8B (Ours) }&0.8B&1024&N@10&57.9&48.8&66.7&45.3&61.7&38.0&75.6&78.1&59.0\\
\rowcolor{cyan!15}
\textbf{EvoEmbedding-2B (Ours)}  &2B  & 1024 & N@10&\textbf{69.9}&\underline{58.3}&76.1&\underline{47.8}&\underline{66.0}&\underline{38.6}&\underline{79.1}&\textbf{81.6}&\underline{64.7}\\
\rowcolor{cyan!15}
\textbf{EvoEmbedding-4B (Ours)} & 4B & 1024 & N@10
& \underline{66.0} & \textbf{61.7} & \underline{78.8} & \textbf{49.2} & \textbf{66.9} & \textbf{41.1} & 77.6 & 80.6 & \textbf{65.2} \\
\bottomrule
\end{tabular}
}
\end{table*}

\section{Experiments}

\subsection{Experimental Setup}
\noindent \textbf{Benchmarks.}
We conduct extensive experiments across 10 benchmarks spanning various tasks, domains, and context scales to comprehensively evaluate the effectiveness of EvoEmbedding.
The evaluation is categorized into two primary tracks:
\begin{itemize}
\item \textbf{Retrieval Tasks}: We assess the model's representation capabilities for information retrieval using datasets including ESG-Reports, MLDR, CovidQA, PeerQA, and QASPER~\citep{zhao2026lmeb}. These benchmarks encompass a wide range of diverse domains, e.g., academic papers, biomedical articles and long-documents.
Alongside these, we evaluate the model on conversational memory datasets, namely REALTALK~\citep{lee2025realtalk}, LoCoMo~\citep{maharana2024evaluating}, and LongMemEval~\citep{wu2024longmemeval}.
\item \textbf{Generation Tasks}:
We evaluate the models on long-term personalization and memory tasks to validate the generalization of EvoEmbedding in downstream generative applications.
This includes the aforementioned LoCoMo and LongMemEval, as well as personalization benchmarks such as PersonaMem (32K)~\citep{jiang2025know} and PersonaMME (32K/128K)~\citep{nie2026personavlm}.
\end{itemize}

\noindent \textbf{Baselines.}
We compare EvoEmbedding against three distinct categories of strong baselines:
(1) \textbf{General Embedding Models \& Lexical Retrieval}: These include state-of-the-art dense retrievers across various parameter scales (e.g., the Qwen3-Embedding series, BGE-M3, Multilingual-e5-large, Jina-v5-text-small, KaLM-Embedding-Gemma3, and All-MiniLM-L6-v2)~\citep{zhang2025qwen3,chen2024bge,wang2024multilingual,zhao2025kalm}, as well as the traditional keyword-based retriever, BM25~\citep{robertson2009probabilistic}.
(2) \textbf{Agentic Memory Systems}: We benchmark a standard RAG equipped with EvoEmbedding-4B against several memory-augmented architectures (e.g., Mem0, LightMem, A-Mem, and MemoryOS~\citep{chhikara2025mem0,fang2025lightmem,xu2026mem}) in generation tasks.
(3) \textbf{Retrieval Optimization Strategies}: This encompasses advanced retrieval pipelines, including LLM-based reranking (e.g., Qwen3-Reranker-4B~\citep{zhang2025qwen3}) and multi-turn reasoning-based retrieval~\citep{nie2026personavlm}.

\noindent \textbf{Implementation Details.}
For retrieval-centric tasks, we report standard information retrieval metrics, including Recall@$k$ and Normalized Discounted Cumulative Gain (NDCG@$k$) at various top-$k$ cutoffs~\citep{zhao2026lmeb}.
For generation-centric tasks, we employ Qwen3-30B-A3B as the generation model and GPT-4o-mini as the judge model for automated evaluation, following the protocol in~\citep{fang2025lightmem}. 
By default, the capacity of the latent memory queue is set to $C=512$ tokens, with $K=16$ latent tokens generated per segment step. The final projected dimension for the embedding is set to $D_{emb}=1024$. 
All training and evaluation procedures are conducted on a single server equipped with 8 NVIDIA H800 GPUs.

\subsection{Main Results}

\begin{figure}[!t]
    \centering
    \includegraphics[width=\textwidth]{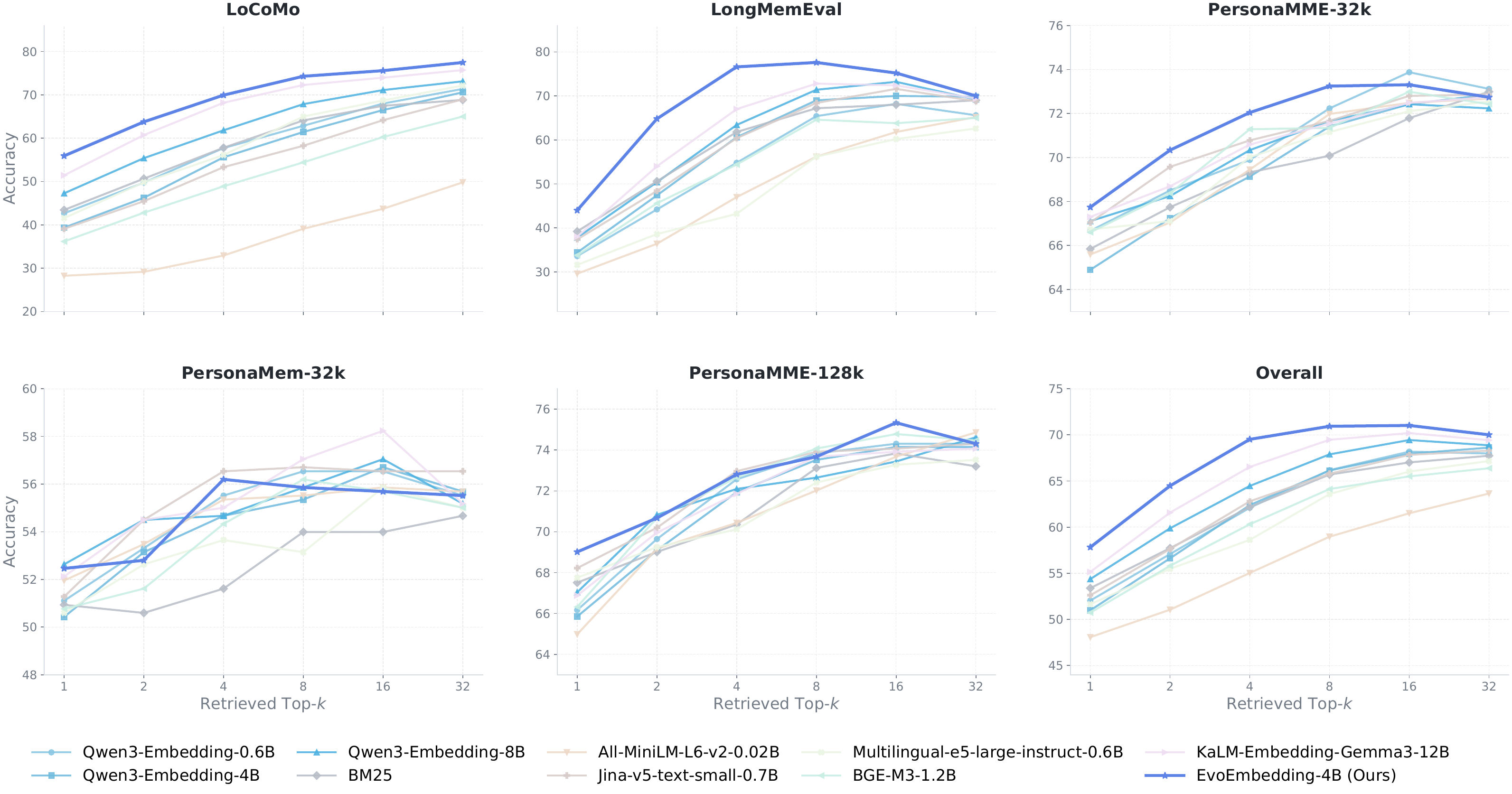}
    \caption{Generation accuracy (\%) of a naive RAG pipeline using different retrieval methods. EvoEmbedding-4B achieves the best overall performance across different retrieval scales.}
    \label{fig:extended_generation}
\end{figure}

\textbf{(a) EvoEmbedding achieves the best overall performance on both retrieval and downstream generation tasks, demonstrating strong scalability and generalization.}
Table~\ref{tab:retrieve_topk_results} and Fig.~\ref{fig:extended_generation} report the retrieval and generation performance of EvoEmbedding across a series of long-context and conversational benchmarks, respectively, where the latter is evaluated using a naive RAG pipeline with top-$k$ retrieved segments as context.

Compared to well-established embedding baselines, our flagship EvoEmbedding-4B establishes the highest Overall Recall@10 (80.5) and NDCG@10 (65.2) on retrieval tasks, surpassing the runner-up KaLM-Embedding-Gemma3~\citep{zhao2025kalm} by substantial absolute margins of 7.8\% and 8.1\%.
Meanwhile, our other variants (0.8B and 2B) also exhibit superior performance against much larger baselines.
On generation tasks, our method consistently achieves the best overall performance across varying Top-$k$ ($k \in \{1, 2, 4, 8, 16, 32\}$) retrieval budgets, as plotted in Fig.~\ref{fig:extended_generation}. Notably, while the accuracy on LoCoMo rises steadily up to $k=32$ without saturation, other datasets generally peak around $k=8$ or $k=16$ and slightly decline thereafter due to the distracting noise introduced by excessively large context windows.
These results demonstrate (1) the \textbf{\textit{effectiveness}} of our approach: utilizing only 180K synthetic samples and standard SFT, EvoEmbedding outperforms peers like KaLM-Embedding-Gemma3 that rely on multi-stage training over 100M data points, highlighting a fundamental architectural advantage in long-context retrieval; and (2) its \textbf{\textit{scalability}} and \textbf{\textit{generalization}}: despite being trained exclusively on shorter samples (maximum $12$K tokens, averaging 1.2K), the model effectively generalizes to 128K-length test scenarios and out-of-domain tasks such as personalized retrieval. This confirms that the high-quality, evolvable representations generated by our model can reliably provide solid grounding for complex retrieval scenarios.

\textbf{(b) Naive RAG with EvoEmbedding surpasses specialized agentic memory systems.} 
Tables~\ref{tab:locomo_results} and~\ref{tab:longmemeval_results} detail the fine-grained performance on the LoCoMo and LongMemEval conversational benchmarks.
Remarkably, a standard naive RAG pipeline powered by EvoEmbedding-4B, utilizing only the retrieved Top-8 segments, consistently surpasses agentic memory architectures (e.g., MemoryOS and Mem0)~\citep{hu2025memory,chhikara2025mem0}.
For instance, on LongMemEval, EvoEmbedding establishes a new state-of-the-art accuracy of 77.6\%, significantly outperforming the best agentic baseline (LightMem, 70.2\%).
Notably, it achieves near-perfect accuracy on the Single-User (98.6\%) and Single-Assistant (100.0\%) subtasks of LongMemEval.
However, on the Temporal and Open-domain subtasks of LoCoMo, EvoEmbedding falls slightly behind the highly specialized LightMem, despite outperforming other static embedding baselines.

Despite this, EvoEmbedding surpasses the full-context baseline on LongMemEval (an absolute gain of +22.8\%) and closely approaches the full-context upper bound on LoCoMo (with a mere 0.6\% gap), which can be further improved to 77.5\% as the retrieval budget increases (Appendix~\ref{app:alg}.3).
Furthermore, as illustrated in Fig.~\ref{fig:intro} (Right), our method resolves the massive token overhead issue typical of agentic systems, incurring \textit{zero} additional token cost since it requires no separate memory construction phase\footnote{The latent memory of EvoEmbedding is constructed during the encoding phase and does not rely on the generator model (i.e., Qwen3-30B-A3B) at test time.}.
This proves that EvoEmbedding can successfully filter out noisy, outdated histories that typically distract the LLM generator.
These results indicate that by embedding temporal and context awareness directly into the representations, our method achieves superior performance in long-context retrieval.

\begin{table}[!t]
\centering
\caption{Evaluation results on LoCoMo. A naive RAG pipeline with EvoEmbedding outperforms agentic memory systems and other embedding baselines.}
\label{tab:locomo_results}
\resizebox{0.9\columnwidth}{!}{
\begin{tabular}{l | cccc | c}
\toprule
\textbf{Method} & \textbf{Single-hop} & \textbf{Multi-hop} & \textbf{Temporal} & \textbf{Open-domain} & \textbf{Overall} \\
\midrule
Full Context & \textbf{87.4} & \underline{69.9} & 51.7 & \underline{57.3} & \textbf{74.9} \\
\midrule
\multicolumn{6}{l}{\textit{Agentic Memory Systems}} \\ \addlinespace[0.1cm]
Mem0      & 67.7 & 54.3 & \underline{57.0} & 46.9 & 61.7 \\
A-MEM            & 67.9 & 57.5 & 27.7 & 43.8 & 56.1 \\
MemoryOS & 72.3 & 62.8 & 33.0 & 51.0 & 61.0 \\
LightMem         & 67.0 & 45.8 & \textbf{76.3} & \textbf{76.8} & 72.6 \\
\midrule
\multicolumn{6}{l}{\textit{NaiveRAG with various embedding models}} \\ \addlinespace[0.1cm]
all-MiniLM-L6-v2         & 29.8 & 20.9 & 40.6 & 49.0 & 39.1 \\
Qwen3-Embedding-8B        & 82.2 & 60.6 & 41.1 & 53.1 & 67.9 \\
KaLM-Embedding-Gemma3-12B  & \underline{87.2} & {62.8} & 47.7 & 52.1 & 72.3 \\
\rowcolor{cyan!15} 
\textbf{EvoEmbedding-4B (Ours)} & 86.6 & \textbf{71.6} & {49.8} & {56.3} & \underline{74.3} \\
\bottomrule
\end{tabular}
}
\end{table}

\begin{table*}[t]
\centering
\caption{
Evaluation results on LongMemEval. A naive RAG pipeline with EvoEmbedding achieves the highest overall accuracy, surpassing both agentic memory systems and the full-context baseline.
The capabilities are evaluated across six categories: Temporal Reasoning (Temp), Multi-Session Dialogue (Multi), Knowledge-Update (Knowledge), Single-User (User), Single-Assistant (Assistant), and Single-Preference (Preference).}
\label{tab:longmemeval_results}
\resizebox{\textwidth}{!}{
\begin{tabular}{l | cccccc | c}
\toprule
\textbf{Model} & \textbf{Temp} & \textbf{Multi} & \textbf{Knowledge}  & \textbf{User} & \textbf{Assistant} & \textbf{Preference} & \textbf{Overall} \\
\midrule
Full Context & 33.1 & 35.6 & 76.9 & 82.9 & 87.5 & 50.0 & 54.8 \\
\midrule
\multicolumn{6}{l}{\textit{Agentic Memory Systems}} \\ \addlinespace[0.1cm]
Mem0             & 41.9 & 28.1 & 28.6 & 55.3 & 26.1 & \textbf{81.8} & 39.5 \\
MemoryOS         & 28.6 & 36.8 & 61.5 & 72.9 & 92.9 & 33.3 & 49.6 \\
A-MEM            & 51.9 & 51.1 & 76.9 & 90.0 & 96.4 & 40.0 & 65.2 \\
LightMem         & 54.2 & 51.9 & 66.7 & 80.0 & 31.3 & \underline{80.0} & 70.2 \\
\midrule
\multicolumn{6}{l}{\textit{NaiveRAG with various embedding models}} \\ \addlinespace[0.1cm]
All-MiniLM-L6-v2     & 40.6 & 34.6 & 70.5 & 77.1 & 96.4 & 60.0 & 56.2 \\
Qwen3-Embedding-8B        & 57.9 & \underline{62.4} & 76.9 & 90.0 & \underline{98.2} & 63.3 & 71.4 \\
KaLM-Embedding-Gemma3-12B   & \underline{60.9} & 58.6 & \underline{80.8} & \underline{92.9} & \underline{98.2} & 73.3 & \underline{72.8} \\
\rowcolor{cyan!15} 
\textbf{EvoEmbedding-4B (Ours)} & \textbf{63.2} & \textbf{71.4} & \textbf{84.6} & \textbf{98.6} & \textbf{100.0} & 60.0 & \textbf{77.6} \\
\bottomrule
\end{tabular}
}
\end{table*}

\begin{table}[!t]
\centering
\caption{Comparison of EvoEmbedding as a plug-and-play module against other strategies (reranking and reasoning) across different agentic memory systems on the LoCoMo benchmark. EvoEmbedding delivers superior overall performance.}
\label{tab:reranker_results}
\resizebox{\columnwidth}{!}{
\begin{tabular}{l l c | cccc | c}
\toprule
\textbf{System} & \textbf{Method} & GPU (G)& \textbf{Single-hop} & \textbf{Multi-hop} & \textbf{Temporal} & \textbf{Open-domain} & \textbf{Overall} \\
\midrule

\multirow{3}{*}{\textbf{A-MEM}} 
& Original & - & 42.91 & 13.71 & 32.29 & 56.48 & 43.57 \\
& $+$ Reasoning & - & 43.26{\color{gray}$_{\scriptscriptstyle +0.4}$} & 15.26{\color{gray}$_{\scriptscriptstyle +1.6}$} & 35.42{\color{gray}$_{\scriptscriptstyle +3.1}$} & 58.98{\color{gray}$_{\scriptscriptstyle +2.5}$} & 45.52{\color{gray}$_{\scriptscriptstyle +2.0}$} \\
& $+$ Qwen3-Reranker-4B   &  14.55    & \underline{61.70}{\color{gray}$_{\scriptscriptstyle +18.8}$} & \underline{22.12}{\color{gray}$_{\scriptscriptstyle +8.4}$} & \underline{43.75}{\color{gray}$_{\scriptscriptstyle +11.5}$} & \underline{76.46}{\color{gray}$_{\scriptscriptstyle +20.0}$} & \underline{60.39}{\color{gray}$_{\scriptscriptstyle +16.8}$} \\
\rowcolor{cyan!10}
& $+$ \textbf{EvoEmbedding-4B (Ours)} & {15.27}   & \textbf{67.38}{\color{gray}$_{\scriptscriptstyle +24.5}$} & \textbf{24.92}{\color{gray}$_{\scriptscriptstyle +11.2}$} & \textbf{45.83}{\color{gray}$_{\scriptscriptstyle +13.5}$} & \textbf{77.53}{\color{gray}$_{\scriptscriptstyle +21.1}$} & \textbf{62.73}{\color{gray}$_{\scriptscriptstyle +19.2}$} \\
\midrule

\multirow{3}{*}{\textbf{LightMem}} 
& Original & -& 41.13 & 23.36 & 41.67 & 46.85 & 40.58 \\
& $+$ Reasoning & - & 48.23{\color{gray}$_{\scriptscriptstyle +7.1}$} & 24.92{\color{gray}$_{\scriptscriptstyle +1.6}$} & 39.58{\color{gray}$_{\scriptscriptstyle -2.1}$} & 49.94{\color{gray}$_{\scriptscriptstyle +3.1}$} & 43.77{\color{gray}$_{\scriptscriptstyle +3.2}$} \\
& $+$ Qwen3-Reranker-4B     & 14.75   & \underline{55.32}{\color{gray}$_{\scriptscriptstyle +14.2}$} & \textbf{29.28}{\color{gray}$_{\scriptscriptstyle +5.9}$} & \underline{44.79}{\color{gray}$_{\scriptscriptstyle +3.1}$} & \underline{61.12}{\color{gray}$_{\scriptscriptstyle +14.3}$} & \underline{52.40}{\color{gray}$_{\scriptscriptstyle +11.8}$} \\
\rowcolor{cyan!10}
& $+$ \textbf{EvoEmbedding-4B (Ours)} & {14.43}    & \textbf{58.51}{\color{gray}$_{\scriptscriptstyle +17.4}$} & \underline{28.66}{\color{gray}$_{\scriptscriptstyle +5.3}$} & \textbf{46.88}{\color{gray}$_{\scriptscriptstyle +5.2}$} & \textbf{63.14}{\color{gray}$_{\scriptscriptstyle +16.3}$} & \textbf{54.09}{\color{gray}$_{\scriptscriptstyle +13.5}$} \\
\midrule

\multirow{3}{*}{\textbf{MemoryOS}} 
& Original & -& 49.65 & 21.81 & 43.75 & 59.10 & 48.64 \\
& $+$ Reasoning & - & 60.64{\color{gray}$_{\scriptscriptstyle +11.0}$} & 28.04{\color{gray}$_{\scriptscriptstyle +6.2}$} & 47.92{\color{gray}$_{\scriptscriptstyle +4.2}$} & 71.94{\color{gray}$_{\scriptscriptstyle +12.8}$} & 59.22{\color{gray}$_{\scriptscriptstyle +10.6}$} \\
& $+$ Qwen3-Reranker-4B   &   19.53   & \underline{64.18}{\color{gray}$_{\scriptscriptstyle +14.5}$} & \underline{38.01}{\color{gray}$_{\scriptscriptstyle +16.2}$} & \underline{52.08}{\color{gray}$_{\scriptscriptstyle +8.3}$} & \textbf{83.47}{\color{gray}$_{\scriptscriptstyle +24.4}$} & \underline{68.51}{\color{gray}$_{\scriptscriptstyle +19.9}$} \\
\rowcolor{cyan!10}
& $+$ \textbf{EvoEmbedding-4B (Ours)} &{16.27}   & \textbf{68.09}{\color{gray}$_{\scriptscriptstyle +18.4}$} & \textbf{38.94}{\color{gray}$_{\scriptscriptstyle +17.1}$} & \textbf{57.29}{\color{gray}$_{\scriptscriptstyle +13.5}$} & \underline{82.28}{\color{gray}$_{\scriptscriptstyle +23.2}$} & \textbf{69.09}{\color{gray}$_{\scriptscriptstyle +20.5}$} \\

\bottomrule
\end{tabular}
}
\end{table}

\begin{figure}[!t]
\centering
\includegraphics[width=1.0\linewidth]{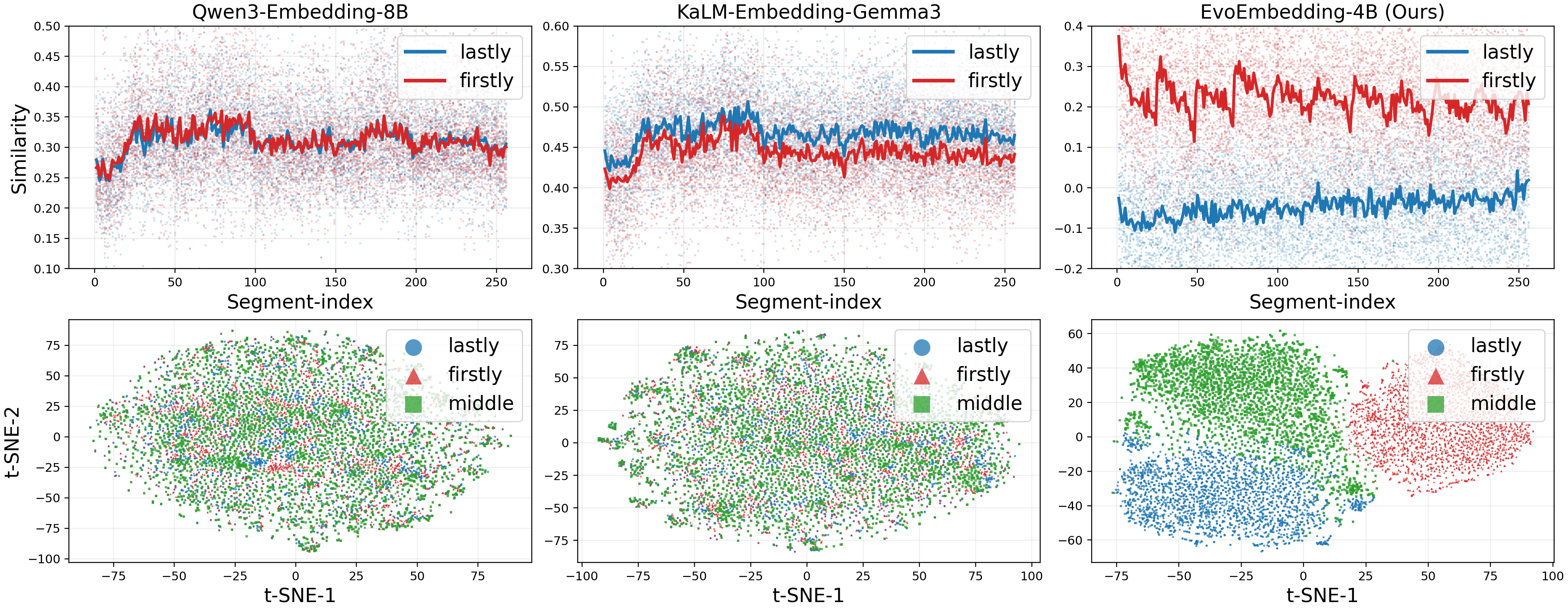}
\caption{\textbf{Analysis of Temporal Query Sensitivity in Long-Context Retrieval.} 
Given 64 randomly selected long-context test samples (each with 256 segments), we query the segments using the template: \textit{``What did I mention [keyword] in our conversation?''} under three temporal settings: ``firstly'', ``lastly'', and ``in the middle''. 
\textbf{(Top Row):} Average similarity curves between queries and historical segments. For EvoEmbedding, the similarity between the query and segments exhibits a chronological increase for the keyword ``lastly'', and peaks sharply at the initial segments for the keyword ``firstly''. This indicates that EvoEmbedding can accurately retrieve context information from both the beginning and the end.
\textbf{(Bottom Row):} t-SNE visualization of query-conditioned segment representations (Hadamard product). Baseline representations remain fully entangled. In contrast, EvoEmbedding's latent space is highly sensitive to temporal semantics.}
\label{fig:vis}
\end{figure}

\textbf{(c) EvoEmbedding enhances existing memory systems via plug-and-play integration.} 
We further evaluate EvoEmbedding's compatibility as a plug-and-play enhancement to upgrade existing memory pipelines. We reproduce A-MEM, LightMem, and MemoryOS on the LoCoMo benchmark, using All-MiniLM-L6-v2 as the baseline retriever.
To test the effectiveness of different context enhancement strategies, we restrict the final generator context to a highly constrained Top-20 (resulting in lower baseline scores than originally reported by~\cite{fang2025lightmem}). In this setting, both EvoEmbedding and Qwen3-Reranker-4B~\citep{zhang2025qwen3} are integrated to rerank the initially retrieved 256 segments and select the top 20 segments to form the final context. As a comparative baseline, the reasoning strategy employs a 2-turn thought-and-retrieval context collection process, retrieving 10 non-overlapping segments in each turn. It incurs no additional GPU memory overhead, but requires two extra generation calls.

As shown in Table~\ref{tab:reranker_results}, EvoEmbedding achieves the best overall performance, providing absolute gains over the original baselines (e.g., +19.2\% on A-MEM and +20.5\% on MemoryOS). It consistently outperforms both the reasoning and reranking strategies across all three frameworks.
Moreover, the GPU memory overheads of our model and Qwen3-Reranker-4B (evaluated with a batch size of 64) are comparable.
This is because EvoEmbedding maintains only a fixed-size latent memory to avoid quadratic complexity.
These results demonstrate the strong adaptability of EvoEmbedding.

\textbf{(d) EvoEmbedding captures contextual order and exhibits strong temporal retrieval capabilities.} 
To explicitly validate the model's sensitivity to temporal semantics, we conduct a fine-grained analysis using queries constrained by time-related keywords (e.g., ``firstly'', ``lastly'', and ``in the middle''), as illustrated in Fig.~\ref{fig:vis}.
The top row displays the average similarity curves across segment indices.
Traditional static embeddings, such as Qwen3-Embedding-8B and KaLM-Embedding-Gemma3, exhibit highly entangled and overlapping curves. This indicates that they rely entirely on coarse textual semantics, failing to differentiate the temporal intent of the queries. In sharp contrast, EvoEmbedding successfully decouples these intents: when queried with ``firstly'', the similarity peaks sharply at the initial segments; when queried with ``lastly'', the similarity exhibits a clear chronological increase, peaking towards the end of the context. 

This temporal awareness is further corroborated by the t-SNE visualizations of the query-conditioned segment representations (Fig.~\ref{fig:vis}, bottom row).
While the latent spaces of the baseline models are mixed, EvoEmbedding structures its latent space distinctly. It cleanly separates the representations into distinct, non-overlapping clusters corresponding to their chronological positions.
This visual evidence proves that our continuously updated latent memory successfully captures long-context temporal information and seamlessly integrates it into the final representations.

\begin{table*}[!t]
\centering
\caption{Ablation study of key designs of EvoEmbedding on five benchmarks. We report the total training time (h), accuracy (\%), and the relative performance drop ($\downarrow$) when specific training strategies are removed. The results demonstrate that the latent memory mechanisms are fundamental to model performance, while our proposed batching strategies are critical for training efficiency.
}
\label{tab:ablation_results}
\resizebox{\textwidth}{!}{
\begin{tabular}{l c | ccccc | c}
\toprule
\textbf{Strategy} & \textbf{Time (h)} & \textbf{LoCoMo} & \textbf{LongMemEval} & \textbf{PersonaMem-32K} &\textbf{PersonaMME-32K} & \textbf{PersonaMME-128K} &  \textbf{Overall} \\
\midrule
\rowcolor{cyan!15}
\textbf{EvoEmbedding-4B} & 26.6 & \textbf{69.9} & \textbf{76.6} &\textbf{56.2} & \textbf{72.0} & \underline{72.8} &  \textbf{69.5} \\
\midrule
\quad w/o Memory Queue
& 91.3 & 17.0{\color{gray}$_{\scriptscriptstyle -52.9}$} & 10.0{\color{gray}$_{\scriptscriptstyle -66.6}$} & 46.9{\color{gray}$_{\scriptscriptstyle -9.3}$} & 64.8{\color{gray}$_{\scriptscriptstyle -7.2}$} & 64.3{\color{gray}$_{\scriptscriptstyle -8.5}$} &  40.6{\color{gray}$_{\scriptscriptstyle -28.9}$} \\

\quad w/o Memory Loss
& 27.7  & 15.2{\color{gray}$_{\scriptscriptstyle -54.7}$} & 11.4{\color{gray}$_{\scriptscriptstyle -65.2}$} & 48.9{\color{gray}$_{\scriptscriptstyle -7.3}$} & 65.5{\color{gray}$_{\scriptscriptstyle -6.5}$} & 64.3{\color{gray}$_{\scriptscriptstyle -8.5}$} &  41.1{\color{gray}$_{\scriptscriptstyle -28.4}$} \\

\quad w/o Length-Weighting
& 26.5 & \underline{68.4}{\color{gray}$_{\scriptscriptstyle -1.5}$} & 73.8{\color{gray}$_{\scriptscriptstyle -2.8}$} &\underline{54.5}{\color{gray}$_{\scriptscriptstyle -1.7}$} & \underline{71.6}{\color{gray}$_{\scriptscriptstyle -0.4}$} & \textbf{73.2}{\color{gray}$_{\scriptscriptstyle +0.4}$} &  \underline{68.3}{\color{gray}$_{\scriptscriptstyle -1.2}$} \\

\quad w/o Segment-Batching
& 101.4 & 66.0{\color{gray}$_{\scriptscriptstyle -3.9}$} & \underline{75.0}{\color{gray}$_{\scriptscriptstyle -1.6}$} & 54.3{\color{gray}$_{\scriptscriptstyle -1.9}$} & 71.2{\color{gray}$_{\scriptscriptstyle -0.8}$} & 71.6{\color{gray}$_{\scriptscriptstyle -1.2}$} & 67.6{\color{gray}$_{\scriptscriptstyle -1.9}$} \\
\bottomrule
\end{tabular}
}
\end{table*}

\subsection{Ablation Studies}
\noindent \textbf{Latent Memory and Queue Design.}
Table~\ref{tab:ablation_results} reports the ablation results across five downstream generation benchmarks.
The latent memory queue serves as the core of our model. Removing either the memory queue setting or the memory loss in Eq.~(\ref{eq3_total}) leads to a severe representation collapse. Consequently, we observe a catastrophic performance degradation exceeding 50\% on LoCoMo and LongMemEval. The performance drop on the other three benchmarks is relatively milder (around 8\%) primarily because they are formulated as multiple-choice questions. Crucially, this queue mechanism enables EvoEmbedding to train directly on mixed-length samples without relying on complex curriculum learning~\citep{bulatov2023scaling}. Furthermore, it restricts the generation overhead to only $K=16$ tokens per step instead of updating the full memory capacity ($C=512$), which effectively boosts the training efficiency by 3.4$\times$.

\noindent \textbf{Segment-Batching and Length-Weighting.} 
First, EvoEmbedding employs a segment-batching technique to process multiple consecutive segments simultaneously, significantly accelerating both training and inference. This strategy yields an almost 3.8$\times$ training speedup (reducing time from 101.4 to 26.6 hours) while bringing an overall performance gain of 1.9\% (Table~\ref{tab:ablation_results}). Second, we introduce a length-weighting technique to elegantly balance sample difficulty and context length during optimization. This regularization prevents the model from biasing towards shorter sequences, contributing an additional 1.2\% improvement to the overall accuracy.

\begin{table}[!t]
\centering
\caption{Efficiency and effectiveness comparison between EvoEmbedding and static embedding models. EvoEmbedding obtains the best LongMemEval performance while requiring substantially lower peak GPU memory, though it incurs additional encoding time due to memory construction.}
\label{tab:inference_efficiency}
\renewcommand{\arraystretch}{1.2}
\resizebox{\textwidth}{!}{
\begin{tabular}{lccccccc}
\toprule
\multirow{2}{*}{\textbf{Model}} & \multirow{2}{*}{\textbf{Size}} & \multicolumn{2}{c}{\textbf{Encoding Time (s) $\downarrow$}} & \multirow{2}{*}{\textbf{Peak VRAM (GB) $\downarrow$}} & \multicolumn{2}{c}{\textbf{Best Performance}} \\
\cmidrule(lr){3-4} \cmidrule(lr){6-7}
 & & \textbf{Context (Avg.)} & \textbf{Query (Avg.)} & & \textbf{Top-$k$} & \textbf{Accuracy (\%) $\uparrow$} \\
\midrule
Qwen3-Embedding-4B        & 4B  & \textbf{3.80}  & \textbf{0.026} & 32.3 & $k=16$ & 70.0 \\
Qwen3-Embedding-8B        & 8B  & 5.52  & 0.027 & 43.1 & $k=16$ & 73.2 \\
KaLM-Embedding-Gemma3     & 12B & 9.89  & 0.034 & 69.3 & $k=4$  & 72.8 \\
\rowcolor{cyan!10} \textbf{EvoEmbedding (Ours)} & {4B}  & 22.08 & 0.065 & \textbf{20.9} & \textbf{$k=8$}  & \textbf{77.6} \\
\bottomrule
\end{tabular}}
\end{table}

\subsection{Efficiency Analysis}
We further analyze the inference efficiency of EvoEmbedding. We use LongMemEval as the testbed and compare EvoEmbedding with strong static embedding models, including Qwen3-Embedding (4B and 8B) and KaLM-Embedding-Gemma3-12B.
For a fair comparison, we set the context encoding batch size to 16 for all methods, and report the average encoding time for both context segments and queries, as well as the peak GPU memory usage.
During inference, EvoEmbedding processes the input sequentially and continuously maintains a latent memory queue that tracks the evolving user state, whereas static embedding models encode context segments in parallel.
As shown in Table~\ref{tab:inference_efficiency}, EvoEmbedding trades off context encoding speed for improved accuracy and reduced peak GPU memory usage. Although it requires more time to encode the context, its compact latent memory state substantially lowers memory consumption and leads to the best retrieval performance.

\section{Discussion and Limitations}
\noindent \textbf{Why EvoEmbedding?}
Retrieval without global context is suboptimal.
It is trivial to construct adversarial cases that deceive static embedding or reranking models—for instance, using shallow keyword traps or paraphrased evidence that requires historical grounding to resolve.
To address this, we design a latent memory queue to maintain a finite, rolling contextual state.
This does not require the model to memorize everything, but rather to know how to query past events. This basic intuition directly inspired the design of EvoEmbedding.

\noindent \textbf{Why latent memory for retrieval instead of generation?}
This choice is driven by three key considerations: 
\textit{\textbf{(i) Controllability:}} Equipping LLMs with native or test-time memory is appealing, yet directly modifying their parameters, activations, or prompts often triggers unpredictable behaviors like catastrophic forgetting and hallucinations~\citep{yu2026latent}. 
\textit{\textbf{(ii) Factuality:}} Under limited memory capacity, generative recall struggles to reliably verify whether specific events occurred, whereas retrieval provides a concrete, verifiable record of historical facts~\citep{huang2025survey}.
\textit{\textbf{(iii) Deployability:}} State-of-the-art commercial models are typically accessible only via black-box APIs, precluding any internal state modifications. 
Consequently, leveraging latent memory to construct retrieval representations offers a far more controllable and deployable paradigm.

\noindent \textbf{Why multi-LoRA design?}
The multi-LoRA design unifies memory, retrieval, and generation capabilities within a single general-purpose language model, eliminating the need for specialized embedding models.
Importantly, this capability decoupling  isolates the training dynamics.
By updating only the task-specific LoRA adapters while keeping the backbone frozen, we avoid catastrophic forgetting in the generation process.
This design renders EvoEmbedding highly flexible: it can run efficiently as a standalone local model or serve seamlessly as a plug-and-play retrieval module within broader agent systems.

\noindent \textbf{Limitations.}
Our current framework has certain limitations.
First, constrained by limited computational resources and training data scale, EvoEmbedding may exhibit degraded performance when applied to out-of-domain scenarios.
Second, the current implementation of EvoEmbedding lacks native support for multimodal retrieval.
While the architecture could be extended to incorporate visual or audio modalities based on existing models, managing long-horizon memories in an omni-modal context would require substantially larger queue capacities and more sophisticated dimensional alignment, which we leave for future exploration.

\section{Conclusion}
This paper introduces EvoEmbedding, a novel family of embedding models designed to overcome the limitations of traditional static representations in long-context scenarios and agentic workflows.
By seamlessly integrating a continuously updated latent memory with sequential text encoding, EvoEmbedding moves beyond static semantic search to achieve precise, context-aware matching. To resolve training inefficiency and representation collapse, we propose a memory queue and dynamic segment-batching techniques. Furthermore, we construct \textbf{EvoTrain-180K}, a large-scale dataset tailored to support the training of evolvable embeddings. Extensive evaluations confirm that EvoEmbedding not only establishes the best overall performance across diverse retrieval benchmarks but also successfully integrates into existing retrieval-augmented frameworks to boost their performance. Overall, our work paves a highly scalable and promising path for future long-context information representation.

\bibliographystyle{plainnat}
\bibliography{citation}
\section*{Appendix}
\setcounter{section}{0}
\renewcommand{\thesection}{\Alph{section}}

\section{Statistics of EvoTrain-180K}
\label{app:data}

To provide a comprehensive understanding of the training data used to optimize EvoEmbedding, we present the detailed statistics of the \textbf{EvoTrain-180K} dataset. The dataset comprises a total of 184,137 training instances, meticulously constructed to encapsulate dynamic state transitions and complex temporal reasoning.

\begin{figure}[h]
    \centering
    \begin{minipage}[t]{0.48\textwidth}
        \centering
        \includegraphics[width=\linewidth]{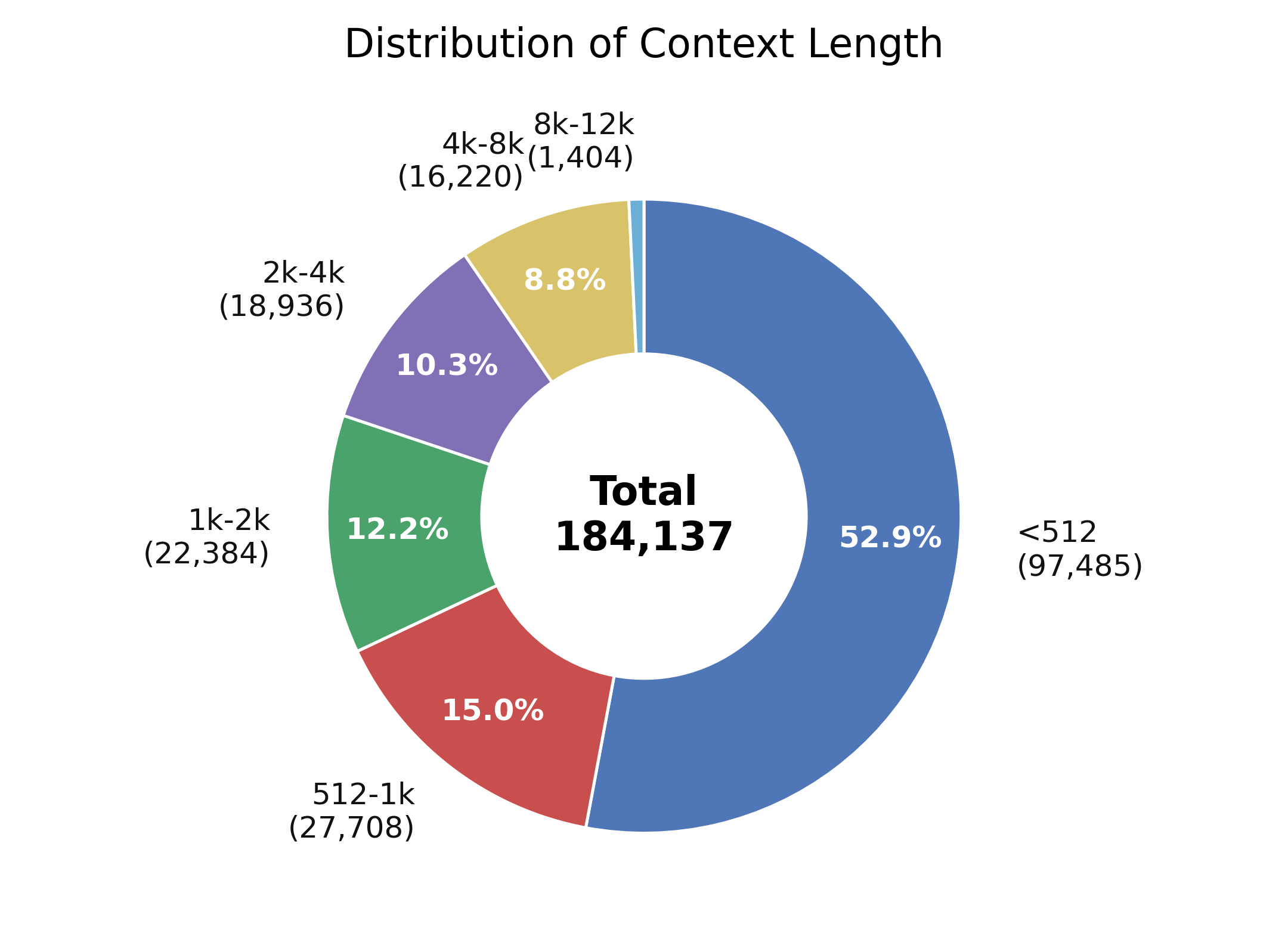}
    \end{minipage}
    \hfill
    \begin{minipage}[t]{0.48\textwidth}
        \centering
        \includegraphics[width=\linewidth]{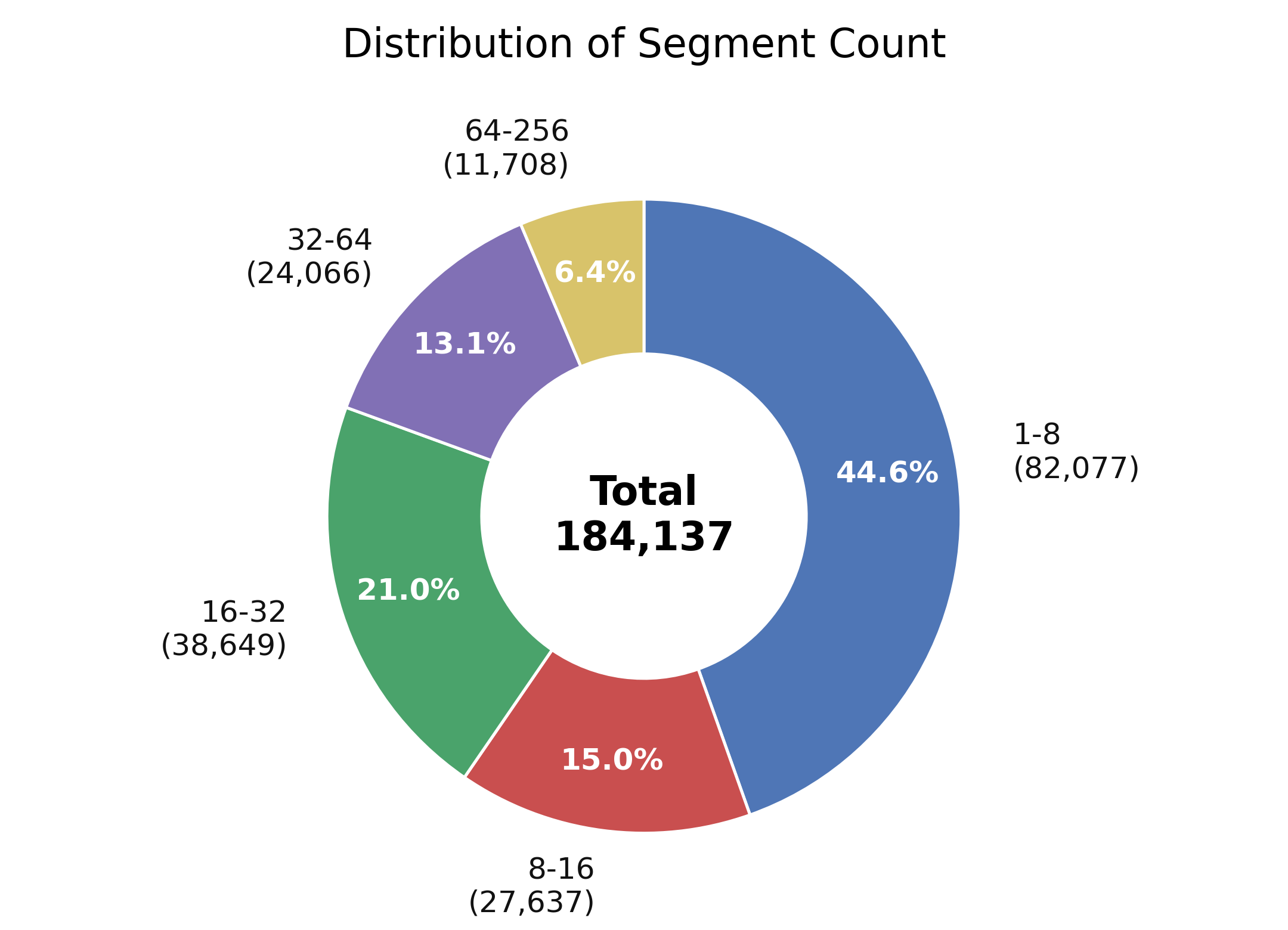}
    \end{minipage}
    \caption{Data distributions of EvoTrain-180K. \textbf{(Left)} The distribution of context lengths (in tokens). \textbf{(Right)} The distribution of segment counts per sample. The highly varied distributions enable the model to learn robust evolvable representations across mixed-length scenarios.}
    \label{fig:data_dist}
\end{figure}

\noindent \textbf{Data Distributions.}
As illustrated in Fig.~\ref{fig:data_dist}, EvoTrain-180K features a highly skewed, mixed-length distribution. While the majority of the contexts (52.9\%) are relatively short (under 512 tokens), there is a significant long-tail distribution extending up to 12K tokens. Similarly, the segment count per sample spans from as few as 2 segments to as many as 246, with 44.6\% of the data falling into the 1-8 segment range. 
This mixed-length nature is crucial: it allows EvoEmbedding to learn basic semantic matching on shorter sequences while mastering temporal dynamics from longer contexts, ultimately enabling it to generalize to 128K testing contexts during inference.

\begin{table}[h]
\centering
\caption{Detailed statistical summary of the EvoTrain-180K dataset.}
\label{tab:data_stats}
\renewcommand{\arraystretch}{1.2}
\begin{tabular}{lrrr}
\toprule
\textbf{Metric} & \textbf{Mean} & \textbf{Min} & \textbf{Max} \\
\midrule
Context Length (tokens) & 1,289.78 & 25 & 10,270  \\
Segment Count per Sample & 20.57 & 2 & 246  \\
Question Length (words) & 15.59 & 3 & 195  \\
Contrastive Negative Samples & 19.45 & 1 & 245  \\
\bottomrule
\end{tabular}
\end{table}

\noindent \textbf{Statistical Summary.} Table~\ref{tab:data_stats} summarizes the key metrics of the dataset. The average context length and segment count are approximately 1.3K tokens and 21, respectively.
Each training instance is paired with an average of 19.45 contrastive negative samples.
Notably, the context lengths and segment counts in our evaluation benchmarks, such as LoCoMo~\citep{maharana2024evaluating}, LongMemEval~\citep{wu2024longmemeval}, and PersonaMME-128k~\citep{nie2026personavlm}, commonly exceed 32K tokens and 500 segments, respectively. Furthermore, the query questions are kept concise (mean length of 15.59 tokens, with 99\% of queries being under 52 tokens), ensuring that the retrieval difficulty stems primarily from the complex temporal contexts rather than convoluted question phrasing.

\section{More Details about EvoEmbedding}
\label{app:alg}

\subsection{Runtime Pipeline}
To provide a clear understanding of the EvoEmbedding architecture, we detail the forward pass processes for both memory evolution and embedding generation.
As illustrated in Algorithms~\ref{alg:memory} and~\ref{alg:embedding}, the two processes share a highly symmetrical architecture, differing primarily in their specific LoRA adapters, appended tokens, and output projection mechanisms. This design allows the model to seamlessly switch between dynamically updating its latent state and generating evolvable embeddings for retrieval.

\begin{figure}[h]
    \centering
    \begin{minipage}[t]{0.49\textwidth}
        \begin{algorithm}[H]
            \caption{Memory Evolution}
            \label{alg:memory}
            \begin{algorithmic}[1]
                \Require Segment $x_t$, queue $\mathbf{M}_{t-1}$
                \State Activate memory LoRA adapter $\theta_m$
                \State $m_{in} \gets \text{LatentProjector}(\mathbf{M}_{t-1})$
                \State $\tilde{x}_t \gets [m_{in} \,;\, x_t \,;\, r_l]$ \hfill $\triangleright$ $r_l$ are $K$ learnable tokens
                \State $\mathbf{\tilde{M}_t} \gets \text{LLM}(\tilde{x}_t)[-K:]$
                \State $\mathbf{M}_t \gets \text{Queue}(\mathbf{M}_{t-1}, f_m(\mathbf{\tilde{M}_t}))$
                \State \Return Updated Memory Queue $\mathbf{M}_t$
            \end{algorithmic}
        \end{algorithm}
    \end{minipage}
    \hfill
    \begin{minipage}[t]{0.49\textwidth}
        \begin{algorithm}[H]
            \caption{Embedding Generation}
            \label{alg:embedding}
            \begin{algorithmic}[1]
                \Require Input $x_t$ (segment/query), queue $\mathbf{M}_{t-1}$
                \State Activate retrieval LoRA adapter $\theta_r$
                \State $m_{in} \gets \text{LatentProjector}(\mathbf{M}_{t-1})$
                \State $\tilde{x}_t \gets [m_{in} \,;\, x_t \,;\, \text{\texttt{<EOS>}}]$
                \State $\mathbf{h_\text{eos}} \gets \text{LLM}(\tilde{x}_t)[-1]$
                \State $\mathbf{v_t} \gets \text{EmbeddingProjector}(\mathbf{h_\text{eos}})$
                \State \Return Representation $\mathbf{v_t}$
            \end{algorithmic}
        \end{algorithm}
    \end{minipage}
\end{figure}

\subsection{Training Details}
For model initialization, EvoEmbedding-0.8B and -2B are derived from Qwen3.5-0.8B and Qwen3.5-2B~\citep{qwen35blog}, respectively, while our flagship EvoEmbedding-4B is initialized with Qwen3-4B~\citep{yang2025qwen3}.
This design ensures a fair comparison against the Qwen3-Embedding baselines, demonstrating that our performance gains stem from the evolving representations rather than from a better base model.
The detailed hyper-parameter configurations used for training are summarized in Table~\ref{tab:hyperparams}. To maximize computational efficiency across highly variable sequence lengths, we enable the length-based grouping strategy (\texttt{group\_by\_length = True}). The entire training process finishes within 11,509 steps, yielding a total training time of 26.6 hours, as reported in Table~\ref{tab:ablation_results}.

\begin{table}[h]
\centering
\caption{Hyper-parameter settings for the training of EvoEmbedding.}
\label{tab:hyperparams}
\renewcommand{\arraystretch}{1.2}
\begin{tabular}{lc}
\toprule
\textbf{Hyper-parameter} & \textbf{Value} \\
\midrule
Learning Rate & $5 \times 10^{-5}$ \\
Batch Size & 16 \\
Training Epochs & 1 \\
Training Steps & 11,509 \\
LR Scheduler & \texttt{cosine} \\
LR Scheduler \texttt{min\_lr} & 0.1 \\
LR Scheduler \texttt{num\_cycles} & 0.5 \\
Group by Length & True \\
\bottomrule
\end{tabular}
\end{table}

\begin{figure}[!t]
    \centering
    \begin{subfigure}[b]{0.48\linewidth}
        \centering
        \includegraphics[width=1.0\linewidth]{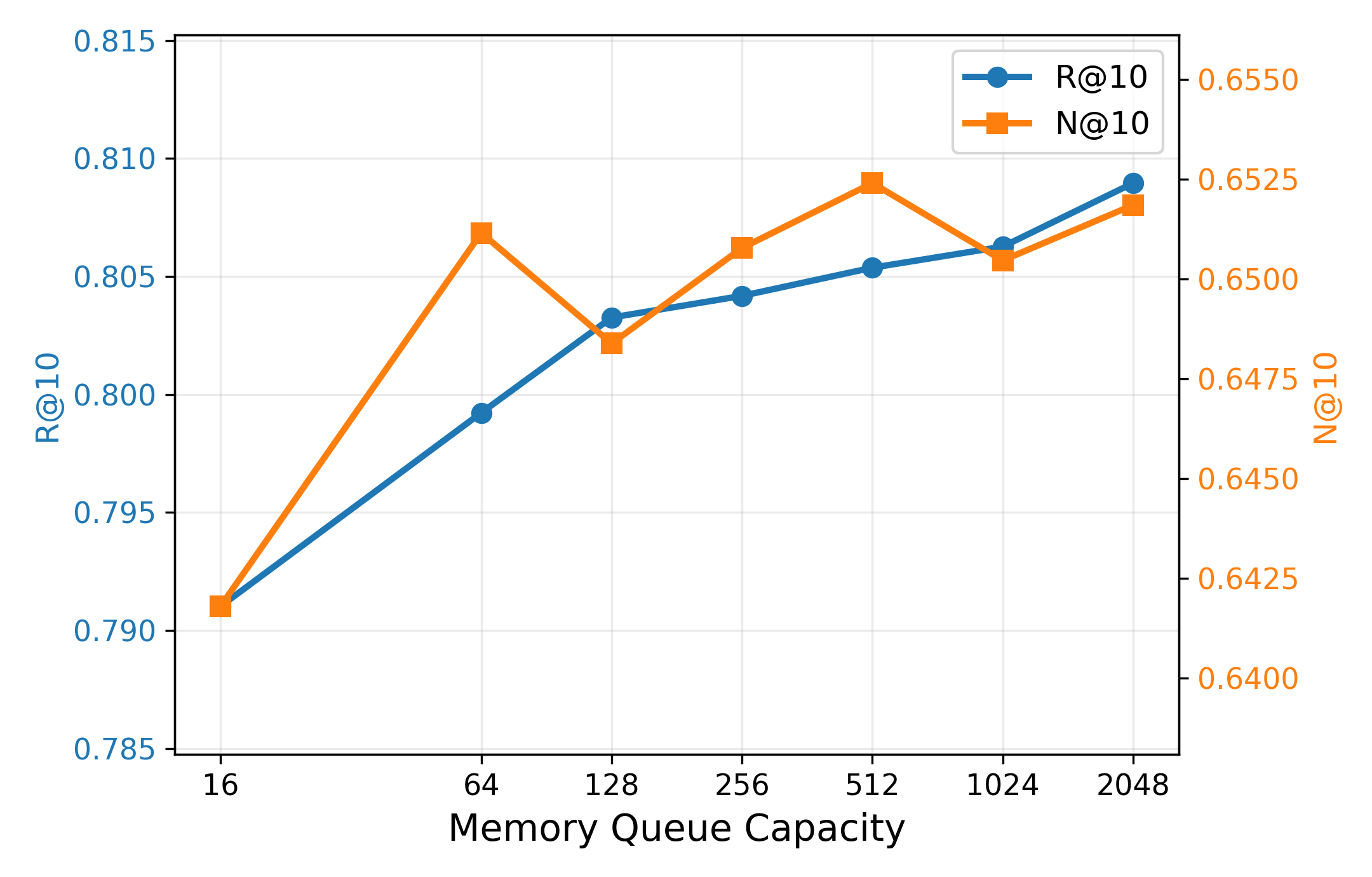}
        \label{fig:sub_1}
    \end{subfigure}
    \hfill 
    \begin{subfigure}[b]{0.48\linewidth}
        \centering
        \includegraphics[width=1.0\linewidth]{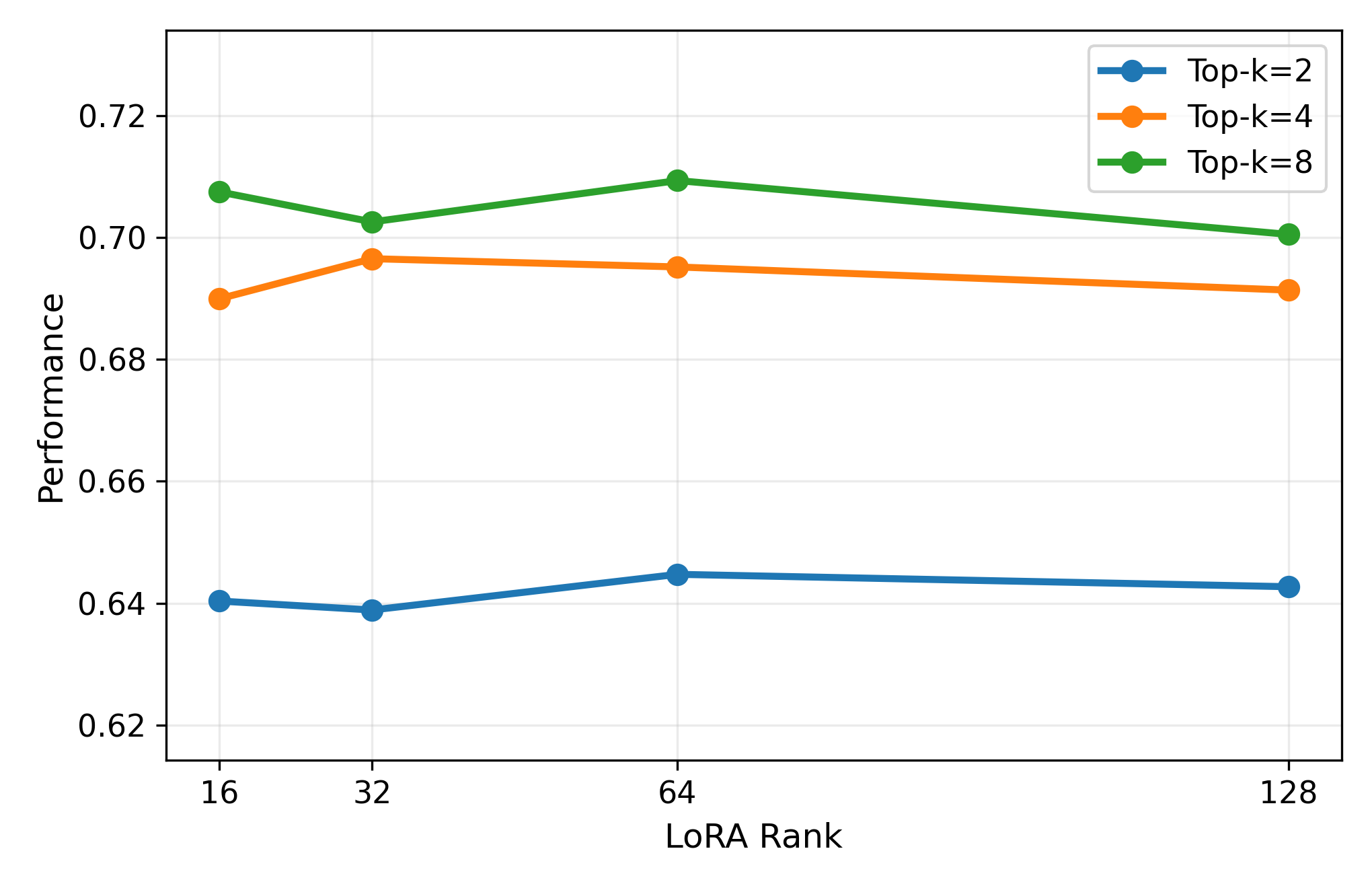}
        \label{fig:sub_2}
    \end{subfigure}
    \caption{Ablation on memory capacity and training scale. \textbf{(Left)} Retrieval performance (Overall R@10 and N@10) across varying memory queue capacities ($C$). A larger queue effectively extends the historical horizon, with performance saturating around $C=512$. \textbf{(Right)} Generation accuracy across varying LoRA Ranks under different Top-$k$ settings. EvoEmbedding demonstrates remarkable parameter efficiency, maintaining highly stable performance even at an extremely low rank ($r=16$).}
    \label{fig:main_vis}
\end{figure}

\subsection{Sensitivity Analysis on Memory Capacity and LoRA Rank}
\noindent \textbf{Memory Capacity.} 
Fig.~\ref{fig:main_vis} (Left) illustrates the retrieval performance across varying memory queue capacities ($C$). As $C$ increases from 16 to 512, we observe a steady upward trend in both R@10 and N@10 metrics. This confirms that a larger queue effectively broadens the historical horizon, allowing the model to better capture long-range temporal dependencies. However, the performance saturates and the NDCG metric peaks when $C$ reaches 512. Beyond this threshold, expanding the queue yields diminishing returns while inevitably increasing memory consumption. Therefore, we set $C=512$ as the default configuration, striking an elegant balance between precise context tracking and computational efficiency.

\noindent \textbf{Impact of LoRA Rank.} 
We further investigate the parameter efficiency by evaluating the generation performance across different LoRA adapter ranks ($r \in \{16, 32, 64, 128\}$).
As shown in Fig.~\ref{fig:main_vis} (Right), the performance curves remain exceptionally stable across all tested ranks and Top-$k$ settings. This indicates that learning evolvable representations does not rely on massive parameter updates; a lightweight adapter is sufficient to activate these representations while fully preserving the original capabilities of the base LLM. We set the final adapter rank to $r=64$ as this configuration achieves a near-optimal peak and ensures maximum representation stability when generalizing to complex tasks.

\subsection{More Experimental Results}

Here, we present the complete results for the generation tasks. Table~\ref{tab:generation_all_topk} details the exact accuracy (\%) of the naive RAG pipeline across five benchmarks under all evaluated retrieval budgets ($k \in \{1, 2, 4, 8, 16\}$). These numerical results confirm EvoEmbedding's consistent superiority across all retrieval budgets.

\begin{table*}[!t]
\centering
\caption{Detailed generation performance (Accuracy \%) across five benchmarks and overall average under varying Top-$k$ retrieved contexts ($k \in \{1, 2, 4, 8, 16\}$). The best results are highlighted in \textbf{bold}, and the second best are \underline{underlined}.}
\label{tab:generation_all_topk}
\resizebox{\textwidth}{!}{
\begin{tabular}{l | c | ccccc | c}
\toprule
\textbf{Model} & \textbf{Top-$k$} & \textbf{LoCoMo} & \textbf{LongMemEval} & \textbf{PersonaMem-32k} & \textbf{PersonaMME-32k} & \textbf{PersonaMME-128k} & \textbf{Overall} \\
\midrule

BM25 & \multirow{12}{*}{$k=1$} & 43.44 & 39.20 & 50.93 & 65.84 & 67.51 & 53.39 \\
All-MiniLM-L6-v2 &  & 28.25 & 29.60 & 51.95 & 65.59 & 64.98 & 48.07 \\
Jina-v5-text-small &  & 39.09 & 37.40 & 51.27 & 67.05 & \underline{68.22} & 52.61 \\
Multilingual-e5-large &  & 41.56 & 31.60 & 50.59 & 66.73 & 67.75 & 51.65 \\
BGE-M3 &  & 36.17 & 33.80 & 50.76 & 66.60 & 66.32 & 50.73 \\
KaLM-Embedding-Gemma3 &  & \underline{51.43} & 38.00 & 52.12 & 67.30 & 66.88 & \underline{55.15} \\
Qwen3-Embedding-0.6B &  & 42.66 & 33.60 & 51.10 & 66.67 & 66.17 & 52.04 \\
Qwen3-Embedding-4B &  & 39.35 & 34.40 & 50.42 & 64.90 & 65.85 & 50.98 \\
Qwen3-Embedding-8B &  & 47.27 & 37.80 & \textbf{52.63} & 67.11 & 67.04 & 54.37 \\
\rowcolor{cyan!15}
\textbf{EvoEmbedding-0.8B (Ours)} &  & 42.40 & 40.20 & 51.61 & \underline{67.49} & 67.83 & 53.91 \\
\rowcolor{cyan!15}
\textbf{EvoEmbedding-2B (Ours)} &  & 47.73 & \underline{41.80} & 50.59 & 67.36 & 66.48 & 54.79 \\
\rowcolor{cyan!15}
\textbf{EvoEmbedding-4B (Ours)} &  & \textbf{55.91} & \textbf{44.00} & \underline{52.46} & \textbf{67.74} & \textbf{69.01} & \textbf{57.82} \\

\midrule

BM25 & \multirow{12}{*}{$k=2$} & 50.65 & 50.60 & 50.59 & 67.74 & 69.01 & 57.72 \\
All-MiniLM-L6-v2 &  & 29.16 & 36.40 & \underline{53.48} & 67.05 & 69.17 & 51.05 \\
Jina-v5-text-small &  & 45.45 & 48.40 & \textbf{54.50} & 69.58 & 70.20 & 57.63 \\
Multilingual-e5-large &  & 49.81 & 38.60 & 52.63 & 67.11 & 69.25 & 55.48 \\
BGE-M3 &  & 42.79 & 45.60 & 51.61 & 68.37 & \underline{70.67} & 55.81 \\
KaLM-Embedding-Gemma3 &  & \underline{60.71} & 54.00 & \textbf{54.50} & 68.69 & 69.96 & 61.57 \\
Qwen3-Embedding-0.6B &  & 49.74 & 44.20 & 53.31 & 68.50 & 69.64 & 57.08 \\
Qwen3-Embedding-4B &  & 46.23 & 47.40 & 53.14 & 67.24 & 69.09 & 56.62 \\
Qwen3-Embedding-8B &  & 55.39 & 50.40 & \textbf{54.50} & 68.25 & \textbf{70.83} & 59.87 \\
\rowcolor{cyan!15}
\textbf{EvoEmbedding-0.8B (Ours)} &  & 56.30 & 59.60 & 52.97 & \textbf{71.28} & 69.09 & \underline{61.85} \\
\rowcolor{cyan!15}
\textbf{EvoEmbedding-2B (Ours)} &  & 54.61 & \underline{62.60} & 51.27 & 69.77 & 69.25 & 61.50 \\
\rowcolor{cyan!15}
\textbf{EvoEmbedding-4B (Ours)} &  & \textbf{63.77} & \textbf{64.80} & 52.80 & \underline{70.34} & \underline{70.67} & \textbf{64.47} \\

\midrule

BM25 & \multirow{12}{*}{$k=4$} & 57.79 & 61.80 & 51.61 & 69.32 & 70.36 & 62.18 \\
All-MiniLM-L6-v2 &  & 32.92 & 47.00 & 55.35 & 69.45 & 70.43 & 55.03 \\
Jina-v5-text-small &  & 53.31 & 60.40 & \textbf{56.54} & 70.78 & \textbf{72.96} & 62.80 \\
Multilingual-e5-large &  & 56.10 & 43.20 & 53.65 & 70.02 & 70.12 & 58.62 \\
BGE-M3 &  & 48.90 & 54.40 & 54.33 & 71.28 & 72.65 & 60.31 \\
KaLM-Embedding-Gemma3 &  & \underline{68.18} & 67.00 & 55.01 & 70.59 & 71.86 & 66.53 \\
Qwen3-Embedding-0.6B &  & 57.73 & 54.80 & 55.52 & 69.89 & 72.57 & 62.10 \\
Qwen3-Embedding-4B &  & 55.58 & 60.60 & 54.67 & 69.13 & 71.94 & 62.38 \\
Qwen3-Embedding-8B &  & 61.82 & 63.40 & 54.67 & 70.34 & 72.09 & 64.46 \\
\rowcolor{cyan!15}
\textbf{EvoEmbedding-0.8B (Ours)} &  & 58.31 & 71.40 & 54.67 & \textbf{72.30} & 71.07 & 65.55 \\
\rowcolor{cyan!15}
\textbf{EvoEmbedding-2B (Ours)} &  & 64.48 & \underline{74.00} & 55.01 & 71.47 & 71.70 & \underline{67.33} \\
\rowcolor{cyan!15}
\textbf{EvoEmbedding-4B (Ours)} &  & \textbf{69.94} & \textbf{76.60} & \underline{56.20} & \underline{72.04} & \underline{72.81} & \textbf{69.52} \\

\midrule

BM25 & \multirow{12}{*}{$k=8$} & 64.09 & 67.20 & 53.99 & 70.08 & 73.12 & 65.70 \\
All-MiniLM-L6-v2 &  & 39.09 & 56.20 & 55.52 & 71.98 & 72.02 & 58.96 \\
Jina-v5-text-small &  & 58.25 & 68.40 & \underline{56.71} & 71.66 & \underline{73.91} & 65.79 \\
Multilingual-e5-large &  & 65.00 & 56.20 & 53.14 & 71.16 & 72.41 & 63.58 \\
BGE-M3 &  & 54.42 & 64.60 & 56.20 & 71.35 & \textbf{74.07} & 64.13 \\
KaLM-Embedding-Gemma3 &  & \underline{72.27} & 72.80 & \textbf{57.05} & 71.54 & 73.68 & 69.47 \\
Qwen3-Embedding-0.6B &  & 62.86 & 65.40 & 56.54 & \underline{72.23} & 73.83 & 66.17 \\
Qwen3-Embedding-4B &  & 61.43 & 69.20 & 55.35 & 71.41 & 73.52 & 66.18 \\
Qwen3-Embedding-8B &  & 67.86 & 71.40 & 55.86 & 71.66 & 72.65 & 67.89 \\
\rowcolor{cyan!15}
\textbf{EvoEmbedding-0.8B (Ours)} &  & 55.71 & 74.80 & 55.52 & 71.92 & 73.44 & 66.28 \\
\rowcolor{cyan!15}
\textbf{EvoEmbedding-2B (Ours)} &  & 69.94 & \underline{76.40} & \underline{56.71} & 72.11 & 73.60 & \underline{69.75} \\
\rowcolor{cyan!15}
\textbf{EvoEmbedding-4B (Ours)} &  & \textbf{74.29} & \textbf{77.60} & 55.86 & \textbf{73.24} & 73.68 & \textbf{70.93} \\

\midrule

BM25 & \multirow{12}{*}{$k=16$} & 67.47 & 68.00 & 53.99 & 71.79 & 73.83 & 67.02 \\
All-MiniLM-L6-v2 &  & 43.70 & 61.80 & 55.86 & 72.49 & 73.68 & 61.50 \\
Jina-v5-text-small &  & 64.16 & 71.60 & 56.54 & 72.80 & 74.07 & 67.83 \\
Multilingual-e5-large &  & 68.70 & 60.20 & 55.86 & 72.11 & 73.28 & 66.03 \\
BGE-M3 &  & 60.26 & 63.80 & 55.69 & 72.99 & \underline{74.78} & 65.50 \\
KaLM-Embedding-Gemma3 &  & 73.96 & 72.40 & \textbf{58.23} & 72.49 & 73.91 & \underline{70.20} \\
Qwen3-Embedding-0.6B &  & 67.99 & 68.20 & 56.54 & \textbf{73.88} & 74.31 & 68.18 \\
Qwen3-Embedding-4B &  & 66.49 & 70.00 & 56.71 & 72.42 & 74.15 & 67.95 \\
Qwen3-Embedding-8B &  & 71.10 & 73.20 & \underline{57.05} & 72.42 & 73.44 & 69.44 \\
\rowcolor{cyan!15}
\textbf{EvoEmbedding-0.8B (Ours)} &  & 59.42 & \underline{74.20} & \underline{57.05} & 72.68 & 74.23 & 67.51 \\
\rowcolor{cyan!15}
\textbf{EvoEmbedding-2B (Ours)} &  & \underline{74.48} & 73.20 & 55.69 & 73.18 & 73.83 & 70.08 \\
\rowcolor{cyan!15}
\textbf{EvoEmbedding-4B (Ours)} &  & \textbf{75.58} & \textbf{75.20} & 55.69 & \underline{73.31} & \textbf{75.34} & \textbf{71.02} \\

\bottomrule
\end{tabular}
}
\end{table*}

\end{document}